\documentclass[runningheads]{llncs}

\usepackage{eccvabbrv}
\usepackage{eccv} 

\usepackage[utf8]{inputenc}
\usepackage[T1]{fontenc}
\usepackage{graphicx}
\graphicspath{{./}{./img/}}
\usepackage{booktabs}
\usepackage{array}
\usepackage{multirow}
\usepackage{adjustbox}
\usepackage{ragged2e}
\usepackage{amsmath}
\usepackage{amssymb}
\usepackage{float}
\usepackage{multicol}
\usepackage{afterpage}
\usepackage{placeins}
\usepackage{caption}
\usepackage[hang,flushmargin]{footmisc}

\DeclareUnicodeCharacter{FF0C}{,}

\usepackage[table]{xcolor}
\definecolor{cvprblue}{rgb}{0.21,0.49,0.74} 
\definecolor{eccvblue}{rgb}{0.21,0.49,0.74} 

\usepackage[breaklinks,colorlinks,citecolor=eccvblue,linkcolor=eccvblue]{hyperref}
\usepackage[capitalize]{cleveref}
\providecommand{\thetitle}{}
\let\titleold\title
\renewcommand{\title}[1]{\titleold{#1}\gdef\thetitle{#1}}

\begin{document}

% ---------------------------------------------------------------
% Title and Authors
% ---------------------------------------------------------------
% \title{Bridging Species and Domains: Semantic Consistency Learning for Generalizable Animal Re-identification}
\title{Cross-Species Animal Re-Identification with Semantic Consistency Learning}

% Short title for running heads
\titlerunning{Cross-Species Animal Re-Identification}

% % Authors
% Second Author\inst{2}\orcidID{1111-2222-3333-4444}}

% % Short author names for running heads

% % Institutes
% \institute{Institution1, Institution1 address \\
% \email{firstauthor@i1.org} \and
% Institution2, Institution2 address\\
% \email{secondauthor@i2.org}}

% Authors
\author{Shuoyi Chen\textsuperscript{*} \and
Yuejia Li\textsuperscript{*} \and
Mang Ye\textsuperscript{\dag}}

% Short author names for running heads
\authorrunning{S. Chen et al.}

% Institutes
\institute{School of Computer Science, Wuhan University, Wuhan, China\
\email{\{chenshuoyi,liyuejia,yemang\}@whu.edu.cn}}

\maketitle

\renewcommand{\thefootnote}{}
\footnotetext{\textsuperscript{*} Equal contribution. \quad \textsuperscript{\dag} Corresponding author.}

% ---------------------------------------------------------------
% Abstract
% ---------------------------------------------------------------
\begin{abstract}
% Animal Re-Identification (ReID) aims to distinguish individual animals across diverse viewpoints and habitats, supporting wildlife monitoring and conservation.
% Unlike person or vehicle ReID conducted in structured settings, animal ReID involves inherently diverse species with distinct morphologies and ecological contexts.
% Collecting sufficient labeled data for every species is impractical, creating a critical need for models that can generalize effectively to unseen species and environments. In this paper, we propose the Semantic Consistency Learning (SCL) framework for cross-species generalization in animal ReID. To stabilize representation learning under environmental and species-induced domain shifts, we design Hierarchical Frequency Distillation, which integrates frequency-domain regularization with temporal knowledge distillation to preserve semantic consistency across domains.
% Furthermore, to bridge fragmented feature spaces across heterogeneous species, we introduce Cross-species Neighborhood Modeling, which learns transferable relational structures through dynamic intra- and inter-species associations.
% Comprehensive evaluations on 58 unseen domains covering more than 40 animal species validate the superior generalization and scalability of SCL, significantly outperforming state-of-the-art approaches and demonstrating its effectiveness as a universal framework for cross-species ReID.

Generalizable animal Re-Identification (ReID) aims to recognize individual animals across species with diverse morphologies and ecological contexts. Unlike person ReID, where different domains share similar body structures, animal species often exhibit drastically different anatomical structures and visual patterns, making it difficult to establish shared visual correspondences. As a result, representations learned across species tend to form fragmented embedding spaces, which severely limits cross-species generalization.
To address this challenge, we propose Semantic Consistency Learning (SCL), a framework designed to learn representations that remain stable across appearance variations while preserving semantic structures shared across species. SCL consists of two complementary components. Foreground--Background Decoupled Spectral Normalization (FDSNorm) stabilizes feature statistics by suppressing environment-induced style variations in a region-aware manner, while Cross-species Neighborhood Modeling (CNM) captures transferable relational structures across species through dynamic feature neighborhoods.
Extensive experiments on 11 public animal ReID datasets demonstrate that SCL consistently outperforms state-of-the-art methods under multiple cross-species evaluation protocols and generalizes effectively to previously unseen species and ecological domains. Code is available at \url{https://github.com/Kemalau/ECCV-26-SCL}.

\end{abstract}

% ---------------------------------------------------------------
% Content
% ---------------------------------------------------------------

\section{Introduction}
\label{sec:intro} 

Animal re-identification (ReID) aims to recognize the same animal individual across different times, viewpoints, and environmental conditions \cite{ye2024transformer, jiao2023toward}. It plays a critical role in ecological monitoring, wildlife conservation, and long-term population analysis. While remarkable progress has been achieved in person and vehicle ReID \cite{chen2022rotation, chen2017person, lou2019veri}, these tasks are typically studied in relatively structured environments where objects share consistent geometric layouts and appearance statistics. In contrast, animal ReID operates in open-world scenarios that involve diverse species, complex habitats, and highly varying visual characteristics. Animals from different species exhibit substantial variations in body morphology, texture patterns, and environmental context. These factors introduce severe distribution shifts that go far beyond the viewpoint or illumination changes commonly addressed in human-centered ReID tasks. As a result, developing models that can generalize across species remains a fundamental challenge.

Most existing animal ReID studies focus on a single-species setting, where models are trained and evaluated using data from one specific species \cite{li2021atrw,korschens2019elpephants,nepovinnykh2022sealid, papafitsoros2022seaturtleid}. Under this formulation, models can learn discriminative representations tailored to the appearance statistics of that species and achieve strong performance on curated benchmarks. However, such a paradigm limits model reusability in real-world applications. Each new species often requires collecting additional data, annotating identities, and retraining models. To address this limitation, recent work has begun to explore multi-species animal ReID, which aims to train a unified model using data from multiple species and generalize to unseen species or new ecological domains. Some approaches attempt to improve robustness by constructing larger and more diverse datasets \cite{hou2024openanimals, vcermak2024wildlifedatasets, jiao2023toward}, while others focus on extracting richer visual cues such as local discriminative patterns \cite{nepovinnykh2024species} or high-frequency information \cite{li2024adaptive}. Although these methods improve recognition accuracy on known species, they mainly enhance instance-level discrimination within species and do not explicitly model how representations should be shared across species. Consequently, the learned embeddings often fail to generalize to unseen species or ecological domains.

\begin{figure}[t]
    \centering
    \includegraphics[width=\linewidth]{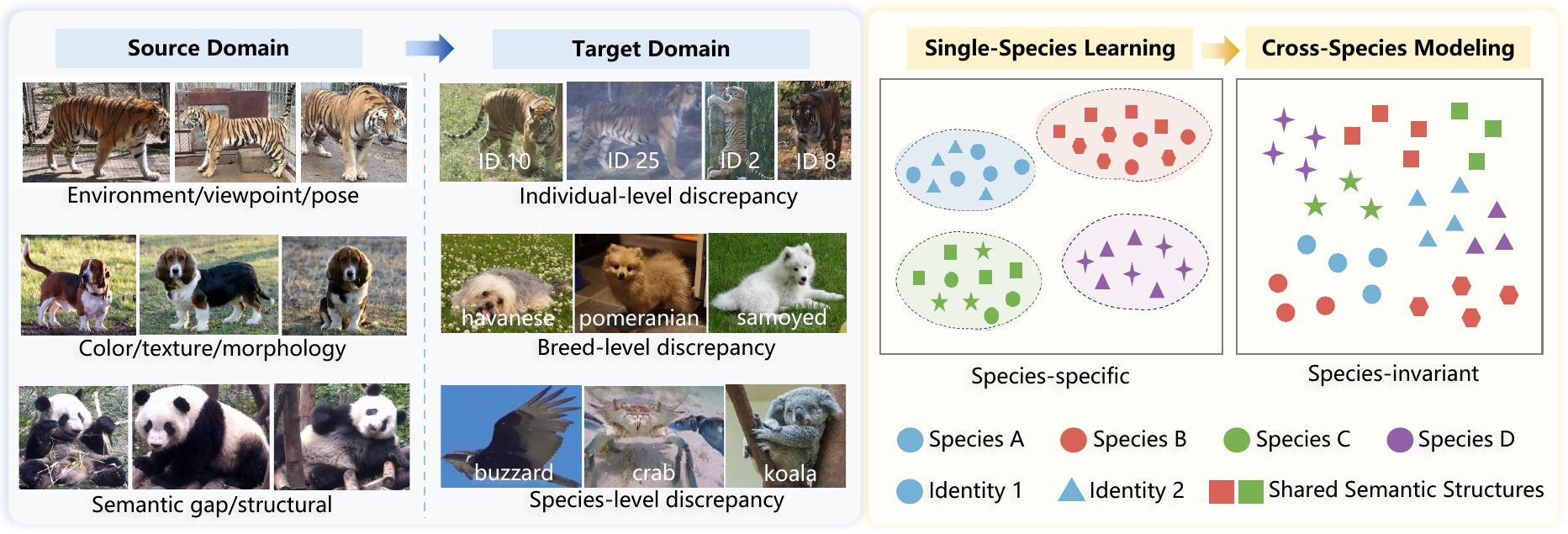}
    \caption{Multi-level domain discrepancies in animal ReID, including individual, breed, and species level variations beyond viewpoint and pose changes.}
    \label{intro}
\end{figure}

% Domain generalization (DG) in person ReID has been widely explored via domain-invariant representation learning \cite{zhang2022learning}, using techniques such as feature disentanglement \cite{zou2020joint, lin2021domain}, style normalization \cite{nie2024rethinking, jin2020style}, or meta-learning \cite{choi2021meta, ni2022meta}. These methods effectively address appearance variations across viewpoints, poses, and illumination conditions. However, while individual-level discrepancies resemble those in person ReID, species-level variations are unique to animals. As shown in \Cref{intro}, breeds differ in color, texture, and morphology, while cross-species discrepancies in structure and semantics create even larger domain gaps. As a result, person DG methods that rely on consistent body structure and shared semantic parts struggle to generalize when applied to animals with diverse appearances and non-uniform anatomical features.

In the person ReID community, domain generalization (DG) has been widely studied to address distribution shifts between source and target domains \cite{lee2025domain, jiang2024domain, Zhao2021M3L}. Existing DG approaches typically aim to learn domain-invariant representations in order to improve cross-domain robustness. Representative techniques include feature disentanglement \cite{zou2020joint,lin2021domain}, style normalization \cite{nie2024rethinking,jin2020style}, and meta-learning \cite{choi2021meta,ni2022meta}. These methods are effective in mitigating variations caused by viewpoint, pose, and illumination changes. However, they implicitly rely on the assumption that objects across domains share consistent semantic structures. For example, human bodies exhibit stable part layouts and similar geometric configurations across domains. This assumption rarely holds in animal ReID. As illustrated in \Cref{intro}, animals present substantial discrepancies at the individual, breed, and species levels. Cross-species variations in body morphology and semantic structure are significantly larger than those encountered in person ReID. Therefore, existing DG methods struggle to learn transferable representations for multi-species animal ReID.

% In this paper, we propose the Semantic Consistency Learning (SCL) framework for cross-species generalization in animal ReID. To stabilize representation learning under environmental and species-induced domain shifts, we design Hierarchical Frequency Distillation (HFD). Specifically, a teacher--student distillation scheme enhanced with hierarchical frequency normalization is employed, where the teacher, updated via exponential moving average (EMA), progressively aggregates long-term semantic knowledge across multiple species. Frequency-domain normalization further regularizes amplitude statistics across layers, suppressing environment-induced feature fluctuations and stabilizing semantic transfer.
% While this improves overall stability, existing ReID methods still emphasize instance discrimination within single-species domains, leading to fragmented representations and poor inter-species generalization.
% To bridge this semantic gap, we introduce Cross-species Neighborhood Modeling (CNM), which jointly enforces intra-species coherence and promotes cross-species generalization.
% By dynamically constructing intra-species and inter-species neighborhoods through mutual-neighbor discovery, this anchors the model to relational topology rather than appearance similarity. This encourages the model to preserve local discriminability while capturing structural regularities shared across species, resulting in a unified and transferable representation space.

These limitations indicate that effective cross-species animal ReID requires representations that remain robust to appearance variations while preserving semantic structures shared across species. To this end, we propose a Semantic Consistency Learning (SCL) framework. The key idea is to stabilize appearance statistics while maintaining structural information in the learned representations.
Specifically, to mitigate representation instability caused by environmental and species variations, we introduce Foreground--Background Decoupled Spectral Normalization (FDSNorm), a frequency-domain normalization mechanism that decouples foreground and background regions for adaptive spectral modulation. Unlike existing normalization strategies that suppress style variations globally in spatial or spectral spaces, our approach explicitly accounts for semantic differences across regions. We preserve the original phase information while adaptively modulating amplitude spectra in different semantic regions, enabling structure-preserving style control under cross-species domain shifts.

However, stabilizing feature statistics alone does not explicitly model cross-species relationships in the embedding space. To address this limitation, we introduce Cross-species Neighborhood Modeling (CNM), which captures relational structures across species through dynamic neighborhood construction. Specifically, CNM discovers mutual neighbors to form both intra-species and inter-species neighborhoods, enabling the model to learn from relational topology rather than relying solely on appearance similarity. This mechanism encourages the learning of structural regularities shared across species while preserving discriminative capability within each species.
In summary, our main contributions are as follows:
\begin{itemize}
\item We introduce Foreground--Background Decoupled Spectral Normalization (FDSNorm), a frequency-domain normalization mechanism that decouples foreground and background regions for adaptive spectral modulation. By preserving phase information while modulating amplitude spectra, FDSNorm suppresses appearance variations and stabilizes feature representations under cross-species domain shifts.
\item We propose Cross-species Neighborhood Modeling (CNM), which explicitly captures relational structures across species through mutual neighbor discovery. CNM dynamically constructs intra-species and inter-species neighborhoods, enabling the model to learn shared semantic regularities across species while preserving discriminative capability within each species.
\item We introduce two complementary evaluation protocols for cross-species animal ReID to enable a comprehensive evaluation of generalization to unseen species. Extensive experiments on 11 public datasets demonstrate consistent improvements over competitive state-of-the-art methods.
\end{itemize}

\section{Related Work}

\textbf{Object ReID.} Re-Identification (ReID) has made substantial progress, with a large body of work focused on person and vehicle identification \cite{zhang2024view, yang2024pedestrian, zhu2024seas, chen2026object}. This has produced many powerful methods, from strong convolutional baselines \cite{luo2019bag, wang2018learning} to more recent transformer-based \cite{he2021transreid} and vision-language \cite{li2023clip} architectures. While these general-purpose frameworks are versatile, their standard implementation requires training a separate, species-specific model for each animal category when adapted for animal ReID. Concurrently, a distinct line of research has emerged that focuses specifically on the challenges of animal ReID. These methods are tailored to specific intra-species challenges, such as identifying livestock by coat patterns \cite{andrew2021friesian}, re-identifying tigers by their unique stripe patterns \cite{li2021atrw}, using high-frequency supervision for fine-grained details \cite{li2024adaptive}, or addressing pose variation with 3D models \cite{zheng2021deep}. 
These domain-generalizable ReID methods rely on consistent body structures and shared semantic correspondences across domains. In cross-species animal ReID, drastic differences in anatomy and visual patterns break this assumption, limiting their ability to learn transferable representations.

\textbf{Domain Generalized ReID.} DG ReID has a substantial literature, particularly in person ReID \cite{nguyen2024tackling, jiang2024domain}, aiming to improve generalization under changes in scene, viewpoint, and illumination. Various approaches have been explored to achieve this. One line of work focuses on normalization-based methods \cite{jin2020style, choi2021meta}, which suppress camera/style statistics to preserve identity cues. Other strategies include employing Mixture-of-Experts (MoE) \cite{Dai2021RaMoE, xu2022mimic} to structure domain variability. These models typically require predefined experts for known domains, making them ill-suited for unseen species. Methods using memory banks \cite{Song2019DIMN, Liao2020QAConv} stabilize matching across domains, but their efficacy diminishes in animal ReID due to high inter-species similarity and vast intra-species diversity, which can pollute the memory bank. Meta-learning frameworks \cite{Bai2021DMGNet, Zhao2021M3L} have also been proposed to simulate test-time shifts. Their limitation lies in the simulation: the meta-tasks often simulate variations in viewpoint or illumination, failing to prepare the model for the drastic object-type shift encountered when generalizing to a new animal species. More recently, data-driven routes like BAU \cite{cho2024generalizable} emphasize augmentations, and CLIP-based methods \cite{zhao2025cilp, zhao2024clip} leverage vision-language priors. While powerful, standard CLIP priors often capture species-level semantics rather than fine-grained individual identity, requiring significant adaptation. Training-free approaches such as Pose2ID \cite{yuan2025from} leverage pose priors at test time, but their reliance on structured human pose limits applicability to animal ReID. 

% Recognizing this gap, UniReID \cite{jiao2023toward} specifically adapted CLIP-based architectures to tackle domain generalized animal ReID and contributed the large-scale Wildlife71 dataset.

\textbf{Multi-species ReID.} More recently, large-scale animal ReID foundation models such as MegaDescriptor \cite{vcermak2024wildlifedatasets} and MiewID \cite{otarashvili2024miewid} have been proposed, leveraging massive community-curated datasets to train unified embedding networks. UniReID \cite{jiao2023toward} specifically adapted CLIP-based architectures to tackle domain generalized animal ReID and contributed the large-scale Wildlife71 dataset. Community benchmarks such as AnimalCLEF \cite{adam2025animalclef} further promote evaluation of individual animal recognition at scale. 
However, these studies mainly focus on constructing large-scale datasets and adopt conventional instance-level metric learning objectives commonly used in ReID. They do not explicitly design mechanisms for generalization to completely unseen species.
% However, these models still rely on instance-level metric learning objectives without explicit cross-species structural alignment, and their generalization to strictly unseen species remains limited. Overall, existing animal ReID approaches are primarily designed for closed-world settings, lacking the generalization capabilities needed for real-world, open-world ecological applications.

% Although these methods alleviate domain generalization issues to some extent, they still struggle when faced with unseen species. Even methods targeting animal DG like UniReID \cite{jiao2023toward} still struggle to achieve uniform, stable, and strong performance across the dual challenge of highly diverse species and varied environmental domains. 
% To bridge this gap, we propose a cross-species generalizable model and learning framework that explicitly targets object-type variation, achieving robust generalization across diverse species in animal ReID.

\begin{figure*}[t]
    \centering
    \includegraphics[width=\linewidth]{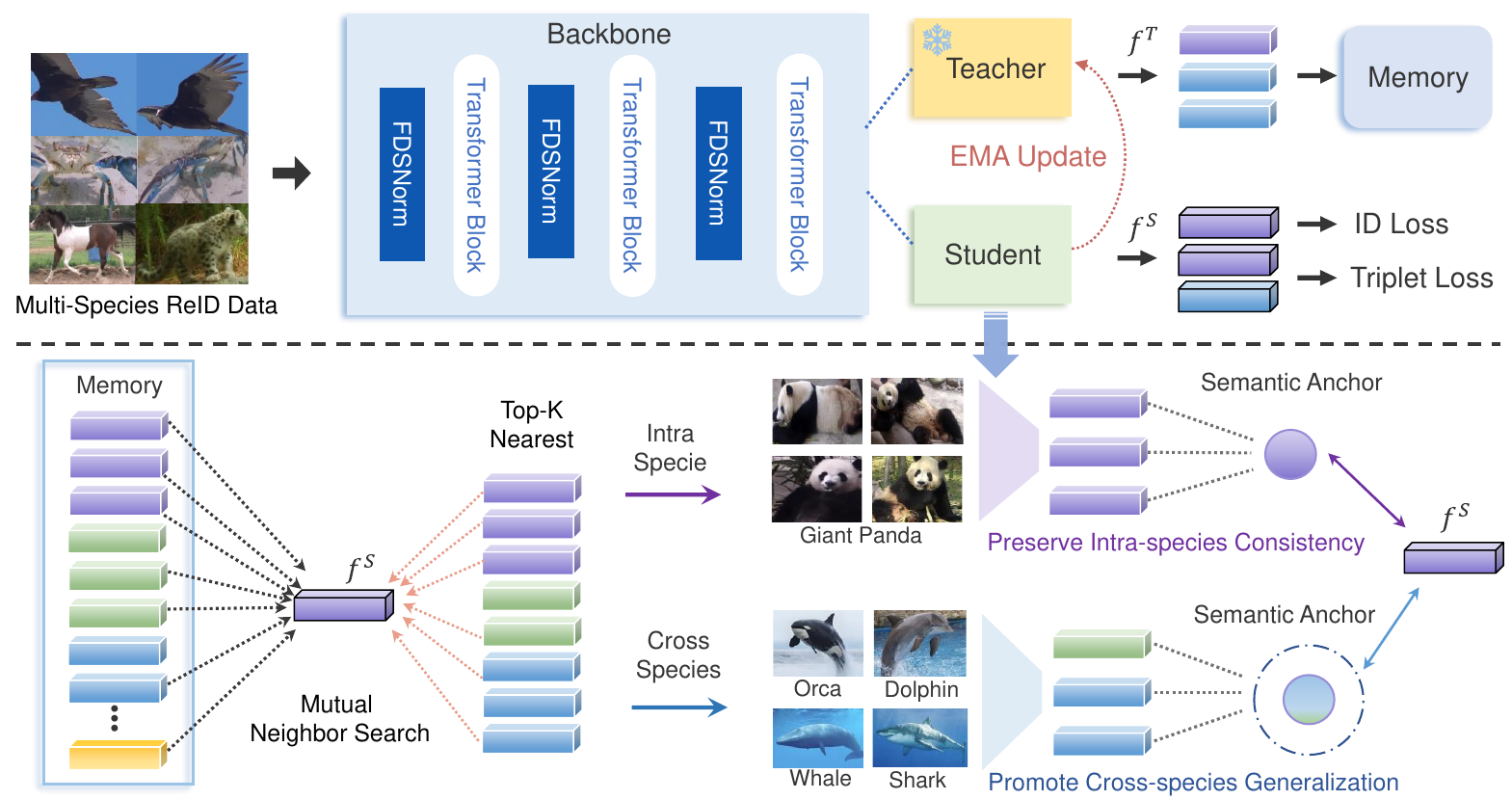}
    \caption{Overview of the proposed Semantic Consistency Learning framework. The upper part illustrates the Foreground--Background Decoupled Spectral Normalization, while the lower part presents Cross-species Neighborhood Modeling.}
    \label{fig:framework}
\end{figure*}

\section{Method}
\subsection{Overview}
The objective of multi-species animal ReID is to learn a unified representation that preserves individual-level discriminability while generalizing across species with diverse morphologies and ecological environments.
However, jointly learning representations from heterogeneous species introduces substantial distributional discrepancies. Differences in texture patterns, body structures, and environmental contexts lead to unstable feature statistics and hinder the formation of a coherent embedding space. As a result, representations learned from different species tend to cluster around species-specific appearance statistics rather than capturing transferable semantic structures.

Furthermore, many existing ReID approaches rely on alignment-based learning strategies that exploit shared visual cues or explicit correspondences across samples. While effective in person ReID or single-species settings, such assumptions rarely hold across species with drastically different anatomies and visual characteristics. Consequently, these methods struggle to establish consistent cross-species representations and often produce fragmented embedding spaces with limited generalization to unseen species.
Domain generalization methods developed for person ReID attempt to mitigate distribution shifts by learning domain-invariant representations through techniques such as feature disentanglement, style normalization, or meta-learning \cite{cho2024generalizable, nie2024rethinking}. However, these approaches implicitly assume comparable semantic structures across domains. In the multi-species setting, where anatomical structures and visual semantics differ substantially, this assumption becomes invalid, limiting their ability to capture transferable representations across species.

To address these challenges, we propose the Semantic Consistency Learning (SCL) framework, which promotes stable and transferable feature learning through two complementary components. As illustrated in \Cref{fig:framework},
(1) Foreground--Background Decoupled Spectral Normalization (FDSNorm) introduces a region-aware frequency-domain normalization mechanism that suppresses environment induced style variations while preserving structure-sensitive semantics.
(2) Cross-species Neighborhood Modeling (CNM) captures relational regularities within and across species by constructing dynamic feature neighborhoods, enabling the model to align transferable semantics while maintaining intra-species discriminative structure.

\subsection{Foreground--Background Decoupled Spectral Normalization}
In multi-species generalized ReID, heterogeneous textures, morphologies, and environmental conditions introduce substantial species-dependent biases, leading to unstable feature distributions. Recent studies show that such variations are closely related to the spectral characteristics of visual representations \cite{lee2023decompose, lin2023deep}. Specifically, the amplitude spectrum mainly captures style-related factors such as illumination and background statistics, whereas the phase spectrum preserves semantic structure. Consequently, normalization strategies that suppress feature statistics uniformly may inadvertently distort phase-dependent semantics.
To address this limitation, we introduce Foreground--Background Decoupled Spectral Normalization (FDSNorm), a frequency-domain normalization mechanism tailored for multi-species ReID. Unlike existing normalization methods that suppress style variations globally in spatial or spectral spaces, FDSNorm explicitly accounts for semantic differences across regions. By preserving the original phase information and adaptively modulating amplitude spectra in foreground and background regions, the proposed mechanism achieves structure-preserving style control under cross-species domain shifts.
% In multi-species generalized ReID, heterogeneous textures, morphologies, and environmental conditions cause domain-specific biases that lead to semantic inconsistency and unstable feature distributions, both in spatial and frequency domains.
% Recent studies have revealed a strong connection between environmental variations and the spectral characteristics of visual representations \cite{lee2023decompose, lin2023deep}.  
% Specifically, the amplitude spectrum predominantly encodes style-related factors such as illumination, color, and background statistics, while the phase spectrum carries semantic and structural information.  
% This implies that conventional normalization methods, which uniformly suppress amplitude variations, may inadvertently distort phase-dependent semantics and weaken representation consistency.  

% \textbf{Frequency-Domain Feature Normalization.}
% Lee et al.~\cite{lee2023decompose} analyzed normalization from a spectral perspective and demonstrated that decoupling amplitude (style) and phase (content) improves domain generalization.  
% Building upon this insight, we extend spectral normalization to Transformer-based backbones and propose hierarchical frequency-domain feature normalization with learnable style control.  
% Unlike convolutional models that operate on spatially structured feature maps, Transformers encode images as sequences of patch tokens, which naturally support hierarchical frequency modeling across layers.  

\textbf{Frequency-Domain Feature Normalization.} Given an input image $\mathbf{I} \in \mathbb{R}^{H_0 \times W_0 \times C_0}$, the Vision Transformer \cite{dosovitskiy2020image} divides it into $N$ non-overlapping patches of size $P \times P$, each projected into a $C$-dimensional embedding space.  
After positional encoding and class-token concatenation, the resulting sequence is processed through $L$ Transformer layers.  
Let $\mathbf{F}^{(l)} \in \mathbb{R}^{B \times C \times H \times W}$ denote the reshaped token feature map at the $l$-th layer, where $B$ is the batch size and $(H, W)$ represent the spatial grid reconstructed from tokens.

To adaptively suppress style-induced domain bias while preserving semantic consistency, we employ a learnable frequency-domain normalization strategy.
First, we obtain a style-normalized version of the feature map via spatial normalization:
\begin{equation}
\tilde{\mathbf{F}}^{(l)} = \text{Norm}(\mathbf{F}^{(l)}),
\end{equation}
where $\text{Norm}(\cdot)$ denotes instance normalization applied across spatial dimensions.
We then perform Discrete Fourier Transform (DFT) on both the original and normalized features:
\begin{equation}
\begin{split}
\mathcal{F}^{(l)}_{\text{org}}(u, v) 
= \sum_{x=0}^{H-1}\sum_{y=0}^{W-1} 
\mathbf{F}^{(l)}(x, y)\, e^{-j 2\pi \left(\frac{ux}{H} + \frac{vy}{W}\right)},
\\
\mathcal{F}^{(l)}_{\text{norm}}(u, v) 
= \sum_{x=0}^{H-1}\sum_{y=0}^{W-1} 
\tilde{\mathbf{F}}^{(l)}(x, y)\, e^{-j 2\pi \left(\frac{ux}{H} + \frac{vy}{W}\right)},
\end{split}
\end{equation}
where $(x,y)$ and $(u,v)$ denote the spatial and frequency coordinates, respectively, and $j = \sqrt{-1}$. 
Each spectral representation is decomposed into amplitude and phase components:
\begin{equation}
\begin{split}
\mathcal{F}^{(l)}_{\text{org}}(u,v) &= A^{(l)}_{\text{org}}(u,v)\, e^{j\Phi^{(l)}_{\text{org}}(u,v)},\\
\mathcal{F}^{(l)}_{\text{norm}}(u,v) &= A^{(l)}_{\text{norm}}(u,v)\, e^{j\Phi^{(l)}_{\text{norm}}(u,v)},
\end{split}
\end{equation}
where $A^{(l)}(u,v) = |\mathcal{F}^{(l)}(u,v)|$ captures the magnitude spectrum and $\Phi^{(l)}(u,v) = \angle \mathcal{F}^{(l)}(u,v)$ encodes structural information.
Following \cite{lee2023decompose}, we treat the amplitude spectrum as a style carrier reflecting environmental variations (\eg, illumination and background), while the phase spectrum represents semantic structure that should remain invariant across domains.

% Instead of global single-strength amplitude mixing, we propose Foreground-Background Decoupled Spectral Normalization (FDSNorm).
Given the CLS token $\mathbf{c}^{(l)}$ and patch tokens $\{\mathbf{t}^{(l)}_i\}_{i=1}^{N}$ at layer $l$, we build a soft foreground mask from CLS-to-patch cosine similarity:
\begin{equation}
s_i=\left\langle \frac{\mathbf{t}^{(l)}_i}{\|\mathbf{t}^{(l)}_i\|},\frac{\mathbf{c}^{(l)}}{\|\mathbf{c}^{(l)}\|}\right\rangle,\quad
\tilde{s}_i=\frac{s_i-\min(\mathbf{s})}{\max(\mathbf{s})-\min(\mathbf{s})+\epsilon},
\end{equation}
\begin{equation}
\tau=\operatorname{Quantile}_{1-r}(\tilde{\mathbf{s}}),\quad
m_i=\sigma\!\left(\frac{\tilde{s}_i-\tau}{T_m}\right),
\end{equation}
where $r$ is the foreground ratio and $T_m$ is the mask temperature. Reshaping $\{m_i\}$ gives $\mathbf{M}^{(l)}\!\in[0,1]^{1\times H\times W}$.
We keep the CLS token unchanged to preserve the global semantic representation, and apply spectral normalization only to patch tokens.

We then apply spatial split before frequency mixing:
\begin{equation}
\mathbf{F}^{(l)}_{\text{fg,org}}=\mathbf{M}^{(l)}\odot\mathbf{F}^{(l)},\quad
\mathbf{F}^{(l)}_{\text{fg,norm}}=\mathbf{M}^{(l)}\odot\tilde{\mathbf{F}}^{(l)},
\end{equation}
\begin{equation}
\mathbf{F}^{(l)}_{\text{bg,org}}=(1-\mathbf{M}^{(l)})\odot\mathbf{F}^{(l)},\quad
\mathbf{F}^{(l)}_{\text{bg,norm}}=(1-\mathbf{M}^{(l)})\odot\tilde{\mathbf{F}}^{(l)}.
\end{equation}
After spatial decoupling, we independently perform DFT on the foreground and background branches to obtain branch-wise amplitude and phase spectra:
\begin{equation}
\mathcal{F}^{(l)}_{\text{fg,*}}=A^{(l)}_{\text{fg,*}}e^{j\Phi^{(l)}_{\text{fg,*}}},\quad
\mathcal{F}^{(l)}_{\text{bg,*}}=A^{(l)}_{\text{bg,*}}e^{j\Phi^{(l)}_{\text{bg,*}}},
\end{equation}
where $* \in \{\text{org}, \text{norm}\}$ denotes original and normalized branches.
Foreground and background use independent mixing strengths:
\begin{equation}
\alpha_{\text{fg}}=\left[\text{softmax}(\boldsymbol{\lambda}_{\text{fg}}/T_s)\right]_0,\quad
\alpha_{\text{bg}}=\left[\text{softmax}(\boldsymbol{\lambda}_{\text{bg}}/T_s)\right]_0,
\end{equation}
where $\boldsymbol{\lambda}_{\text{fg}}, \boldsymbol{\lambda}_{\text{bg}} \in \mathbb{R}^{2}$ are learnable two-dimensional parameter vectors initialized to zeros, $T_s$ is the temperature parameter in softmax, and $[\cdot]_0$ selects the first element as the normalized mixing weight.
\begin{equation}
\hat{A}^{(l)}_{\text{fg}}=\alpha_{\text{fg}}A^{(l)}_{\text{fg,norm}}+(1-\alpha_{\text{fg}})A^{(l)}_{\text{fg,org}},\;
\hat{A}^{(l)}_{\text{bg}}=\alpha_{\text{bg}}A^{(l)}_{\text{bg,norm}}+(1-\alpha_{\text{bg}})A^{(l)}_{\text{bg,org}}.
\end{equation}
In this work, we use the unconstrained variant, i.e., no explicit ordering constraint is imposed between $\alpha_{\text{fg}}$ and $\alpha_{\text{bg}}$, allowing the network to adaptively discover when background regions require stronger style suppression.
Reconstruction preserves branch-wise original phase:
\begin{equation}
\hat{\mathbf{F}}^{(l)}_{\text{fg}}=\mathcal{F}^{-1}\!\left(\hat{A}^{(l)}_{\text{fg}}e^{j\Phi^{(l)}_{\text{fg,org}}}\right),\;
\hat{\mathbf{F}}^{(l)}_{\text{bg}}=\mathcal{F}^{-1}\!\left(\hat{A}^{(l)}_{\text{bg}}e^{j\Phi^{(l)}_{\text{bg,org}}}\right),
\end{equation}
\begin{equation}
\hat{\mathbf{F}}^{(l)}=\hat{\mathbf{F}}^{(l)}_{\text{fg}}+\hat{\mathbf{F}}^{(l)}_{\text{bg}}.
\end{equation}
Through empirical analysis, the FDSNorm module is inserted at multiple Transformer depths ($l \in \{0,4,8\}$), corresponding to the patch-embedding output and intermediate blocks, thereby forming a progressive de-stylization pipeline. Specifically, shallow layers mainly clean pixel-level domain shifts (e.g., illumination and color temperature); middle layers align structure-level shifts (e.g., parts and shape); and deeper layers compensate for residual shifts leaked by residual connections. In parallel, as the CLS token passes through more attention layers, its semantic awareness becomes stronger. This allows the foreground mask to be refined from coarse to fine, yielding a coordinated progression between de-stylization strength and mask quality. In addition, the foreground ratio $r$ and branch-wise mixing strengths ($\alpha_{\text{fg}}$ and $\alpha_{\text{bg}}$) are dynamically learnable parameters, while the mask temperature $T_m$ remains fixed.

\textbf{Temporal Semantic Distillation.}
While normalization suppresses style-related instability within the backbone, temporal inconsistency may still arise from noisy or domain-biased updates during optimization.  
To stabilize the evolution of semantic representations, we employ a teacher--student framework in which the teacher network maintains an exponential moving average (EMA) of the student parameters:
\begin{equation}
\boldsymbol{\theta}_t \leftarrow \mu\,\boldsymbol{\theta}_t + (1-\mu)\,\boldsymbol{\theta}_s,
\end{equation}
where $\boldsymbol{\theta}_t$ and $\boldsymbol{\theta}_s$ denote the teacher and student parameters, respectively, and $\mu \in (0,1)$ is a momentum coefficient.  
The teacher network provides temporally smoothed features that serve as stable semantic anchors for subsequent neighborhood consistency learning, 
effectively distilling long-term structural knowledge into the student without introducing additional supervision.  

\subsection{Cross-species Neighborhood Modeling}
% While Hierarchical Frequency Distillation stabilizes the semantic space, it does not explicitly enforce structural relations among species.
In multi-species ReID, each species forms a visually coherent cluster within the feature space, yet these clusters remain topologically isolated due to the absence of shared semantic anchors. This leads to semantic fragmentation: features are discriminative within species but unaligned across them, hindering transfer to unseen species. Conventional ReID metric learning objectives such as triplet loss \cite{hermans2017defense} rely on instance-level correspondences and fail to exploit the latent relational regularities shared across species.
To bridge this gap, we propose Cross-species Neighborhood Modeling, which constructs dynamic relational structures using a teacher-maintained memory to jointly enforce intra-species consistency and cross-species semantic connectivity.

\textbf{Dynamic Memory Construction.}
Let $\mathbf{f}_t$ and $\mathbf{z}_s$ denote the global features from the teacher and student networks, respectively.
We maintain a feature memory queue 
$\mathcal{M} = \{ (\mathbf{f}_i, sp_i, y_i) \}_{i=1}^{|\mathcal{M}|}$, 
where each entry consists of a normalized teacher feature $\mathbf{f}_i$, its species label $sp_i$, and identity label $y_i$.  
After each iteration, newly computed teacher features are enqueued, while the oldest entries are dequeued to maintain a fixed capacity, ensuring $\mathcal{M}$ captures long-term semantic structure across species.  
For each student feature $\mathbf{z}_s$ with species label $sp$, the cosine similarity to all memory entries is computed as:
\begin{equation}
\mathrm{sim}(\mathbf{z}_s, \mathbf{f}_i) 
= \frac{\mathbf{z}_s^\top \mathbf{f}_i}{\|\mathbf{z}_s\|\|\mathbf{f}_i\|}.
\end{equation}

\textbf{Mutual Neighborhood Search.}
Let $\mathcal{P} = \{\mathbf{f}_t\} \cup \mathcal{M}$ denote the union of the current-batch teacher features and the memory queue.
Each anchor $\mathbf{z}_s$ retrieves two types of neighborhoods from $\mathcal{P}$:
an \textit{intra-species neighborhood} 
$\mathcal{N}_{intra}(\mathbf{z}_s)$ consisting of top-$K_1$ nearest neighbors whose species label matches the anchor, i.e., $sp_i = sp$, 
and a \textit{cross-species neighborhood} 
$\mathcal{N}_{cross}(\mathbf{z}_s)$ of top-$K_1$ nearest neighbors where $sp_i \neq sp$.  
To suppress incidental correlations in the cross-species neighborhood, we adopt a reciprocal nearest-neighbor rule:
for an anchor $A$ and a candidate neighbor $B$ retrieved from the top-$K_1$ cross-species list of $A$, $B$ is retained only if $A$ also appears in the top-$K_2$ cross-species nearest list of $B$ within $\mathcal{P}$.
Here, $K_1$ determines the size of the candidate neighborhood for retrieval, whereas $K_2~(K_2\!\le\!K_1)$ controls the stringency of the reciprocal verification: a larger $K_1$ broadens the candidate pool, while a smaller $K_2$ retains only strongly mutual cross-species pairs. This reciprocal filtering is applied solely to the cross-species neighborhood.
This bidirectional filtering yields more stable semantic neighborhoods that reflect intrinsic relational similarity across species.
The neighborhood centers are defined as:
\begin{equation}
\mathbf{c}_{intra} =
\frac{1}{|\mathcal{N}_{intra}|}
\sum_{\mathbf{f}_i \in \mathcal{N}_{intra}} \mathbf{f}_i,\quad
\mathbf{c}_{cross} =
\frac{1}{|\mathcal{N}_{cross}|}
\sum_{\mathbf{f}_i \in \mathcal{N}_{cross}} \mathbf{f}_i.
\end{equation}
% and all vectors are $\ell_2$-normalized prior to similarity computation.

\textbf{Loss Formulation.}
Cross-species neighborhood modeling jointly optimizes two complementary objectives: 
intra-species compactness and cross-species relational consistency.
\textit{1) Intra-species Compactness.}
For each sample, we align the student feature with its intra-species neighborhood center to enforce intra-species compactness, where the center is computed over same-species neighbors regardless of identity:
\begin{equation}
\mathcal{L}_{\text{intra}} = 1 - \frac{1}{B}\sum_{i=1}^{B} \text{sim}(\mathbf{z}_i, \mathbf{c}_{\text{intra}}^{(i)}),
\end{equation}
where $B$ denotes the batch size and $\text{sim}(\cdot,\cdot)$ denotes cosine similarity.
\textit{2) Cross-species Relational Constraint.}
To bridge gaps across species, we introduce a margin-based relational constraint:
\begin{equation}
\mathcal{L}_{\text{cross}} = \frac{1}{B}\sum_{i=1}^{B} \max\left(0, m - \text{sim}(\mathbf{z}_i, \mathbf{c}_{\text{cross}}^{(i)})\right),
\end{equation}
where $m$ is a margin. Although cross-species centers provide transferable semantic cues, overly strong attraction toward them may pull features of different identities too close, impairing identity discrimination.
The hinge thus acts as a bounded attraction: each feature is pulled toward its cross-species relational center only until the similarity reaches the target margin $m$, after which no gradient is applied. This injects a controlled level of cross-species connectivity (i.e., $\mathrm{sim}\!\ge\!m$), while the small margin and the saturation of the hinge prevent features from being driven into full alignment, thereby preserving identity-level discrimination.
 % encouraging alignment with intra-species centers while maintaining semantic proximity to transferable cross-species clusters.
This formulation softly aligns inter-species manifolds without collapsing structural diversity.
The overall CNM objective is:
\begin{equation}
\mathcal{L}_{CNM} = 
\lambda_{intra}\mathcal{L}_{intra} + 
\lambda_{cross}\mathcal{L}_{cross}.
\end{equation}
In summary, our learning objective is the total loss $\mathcal{L}$, formulated as a weighted sum of the identification loss, the triplet loss, and the CNM loss.
\begin{equation}
\mathcal{L}
= \lambda_{\mathrm{id}} \, \mathcal{L}_{\mathrm{id}}
+ \lambda_{\mathrm{tri}} \, \mathcal{L}_{\mathrm{tri}}
+ \mathcal{L}_{CNM}.
\end{equation}
\textbf{Memory Update.}
After each iteration, normalized teacher features with their species and identity labels are added to the memory:
\begin{equation}
\mathcal{M} \leftarrow 
\mathrm{enqueue}(\mathrm{Norm}(\mathbf{f}_t), sp_t, y_t),
\quad 
\mathrm{dequeue\ oldest}.
\end{equation}
This online update maintains a temporally smoothed and semantically consistent teacher space, providing robust relational guidance for the student network.

\section{Evaluation Protocol}
\subsection{Datasets and Evaluation Protocols}

% We propose a new, unified, and clear evaluation protocol to validate cross-domain and cross-species animal re-identification performance.
% We evaluate our method on 11 publicly available animal ReID datasets on two protocols.

% To facilitate a systematic evaluation of cross-species generalization, we introduce two complementary evaluation protocols for animal ReID. We conduct experiments on 11 publicly available datasets under these protocols to assess both cross-domain and cross-species generalization performance.
\textbf{Datasets and Splits.}
To comprehensively evaluate cross-species generalization, we conduct experiments on 11 publicly available animal ReID datasets covering diverse habitats and species morphologies. This diverse dataset collection provides a challenging evaluation setting with substantial variations in visual appearance and environmental conditions.
\textit{(1) Wildlife71 Dataset} \cite{jiao2023toward}. This large-scale benchmark comprises 71 species. Following its official split protocol, we use the predefined 67 seen species for training.
\textit{(2) PetFace Dataset} \cite{shinoda2025petface}. This dataset contains facial images from 13 domestic animal species, and we test the model separately on each species.
\textit{(3) Nine public datasets}. We further evaluate on iPanda-50 \cite{wang2021giant}, ELPephants \cite{korschens2019elpephants}, SealID \cite{nepovinnykh2022sealid}, GZGC (zebra and giraffe domains) \cite{parham2017animal}, WhaleSharkID \cite{Holmberg_2009}, ATRW \cite{li2021atrw}, HyenaID2022 \cite{wildme_hyenaid2022}, LeopardID2022 \cite{wildme_leopardid2022}, and SeaTurtleID2022 \cite{Adam_2024_WACV}.
As shown in \cref{tab:evaluation_protocols}, we design two complementary evaluation protocols to simulate different cross-species generalization scenarios.
Both protocols enforce a fully open-set setting in which identities and species in the test set are disjoint from those used during training, ensuring that the evaluation reflects genuine cross-species generalization rather than dataset-specific overlap.
Since existing animal ReID datasets adopt inconsistent split strategies or lack official training/testing partitions, we use the full datasets under both protocols to maintain consistent evaluation conditions. The only exception is PetFace, for which we follow the official split. This unified data usage ensures fair and reproducible comparisons across different methods.

\begin{table*}[t]
\centering
\tiny

%==================== (a) Independent table ====================
\begin{minipage}[t]{0.33\textwidth}
\centering
\captionof{table}{Dataset Statistics}
\label{tab:dataset_statistics}

\renewcommand{\arraystretch}{1}
\setlength{\tabcolsep}{3pt}
\begin{tabular}{@{}lccc@{}}
\toprule
Dataset & \#Image & \#ID & Species \\
\midrule
PetFace \cite{shinoda2025petface} & 115,708 & 55,686 & 13 \\
Wildlife71 \cite{jiao2023toward} & 108,096 & 1,924 & 67 \\
iPanda-50 \cite{wang2021giant} & 6,874 & 50 & 1 \\
ELPephants \cite{korschens2019elpephants} & 2,078 & 276 & 1 \\
SealID \cite{nepovinnykh2022sealid} & 2,080 & 61 & 1 \\
GZGC \cite{parham2017animal} & 4,948 & 1,762 & 2 \\
ATRW \cite{li2021atrw} & 2,950 & 135 & 1 \\
HyenaID2022 \cite{wildme_hyenaid2022} & 3,129 & 256 & 1 \\
LeopardID2022 \cite{wildme_leopardid2022} & 6,806 & 430 & 1 \\
SeaTurtleID2022 \cite{Adam_2024_WACV} & 8,729 & 438 & 1 \\
WhaleSharkID \cite{Holmberg_2009} & 7,693 & 543 & 1 \\
\bottomrule
\end{tabular}
\end{minipage}
\hfill
%==================== (b) Independent table ====================
\begin{minipage}[t]{0.59\textwidth}
\centering
\captionof{table}{Evaluation Protocols}
\label{tab:evaluation_protocols}

\tiny
\setlength{\tabcolsep}{0.5pt}
\begin{tabular}{@{}ccc@{}}
\toprule
Protocol & Training Data & Testing Data \\
\midrule
\multirow[c]{5}{*}{1} & \multirow[c]{5}{*}{Wildlife71} & \shortstack{PetFace, iPanda-50} \\
& & \shortstack{ELPephants, SealID,} \\
& & SeaTurtleID2022, ATRW, \\
& & \shortstack{HyenaID2022, LeopardID2022,} \\
& & \shortstack{GZGC, WhaleSharkID,} \\
\midrule
\multirow[c]{5}{*}{2} & \shortstack{iPanda-50+ELPephants} & \multirow[c]{5}{*}{\shortstack{PetFace,\\ Wildlife71}} \\
& \shortstack{+LeopardID2022+GZGC} & \\
& \shortstack{+ATRW+HyenaID2022} & \\
& \shortstack{+SealID+SeaTurtleID2022} & \\
&+WhaleSharkID & \\
\bottomrule
\end{tabular}
\end{minipage}

\end{table*}

\textbf{Evaluation Metrics}. We adopt Cumulative Matching Characteristics (CMC) at Rank-1 and mean Average Precision (mAP) as standard metrics. Since most animal ReID datasets lack explicit camera annotations, we include all valid gallery matches in evaluation rather than only cross-camera ones.

% We additionally report mean Inverse Negative Penalty (mINP) \cite{ye2021deep} to assess performance on the hardest matches, as Rank-k scores alone can be inflated by easy samples.

\section{Experiments}
\subsection{Implementation Details}
All experiments were conducted on four NVIDIA 4090 GPUs using PyTorch. We employ a ViT \cite{dosovitskiy2020image} pre-trained on ImageNet-1K as the backbone; unless a method has specific architectural constraints, all baselines share the same backbone for fair comparison. Input images are resized to $256 \times 256$ with patch size $16 \times 16$. Training augmentations include random horizontal flipping (50\%) and 10-pixel padding. For CNM hyperparameters, we set $K_1{=}8$, $K_2{=}3$, margin${}=0.2$, and memory size${}=4096$. The model is trained for 60 epochs with SGD (initial lr $0.004$, cosine decay) and a total batch size of 128 (8 identities $\times$ 4 images per GPU $\times$ 4 GPUs). Both CNM and FDSNorm use a 3-epoch warm-up. At test time, only original features are used for distance computation.

\setlength{\textfloatsep}{6pt}
\setlength{\dbltextfloatsep}{6pt}

\begin{table*}[!t]
\centering
\caption{Protocol-1 results on 10 unseen domains. The symbol $\dagger$ denotes results obtained from re-implemented versions of the corresponding methods.}

\label{tab:wildlife_complete}
% \scriptsize
\setlength{\tabcolsep}{6pt}
\renewcommand{\arraystretch}{1.10}
\newcommand{\rankonehdr}{\makebox[1.08cm][c]{Rank1}}
\newcommand{\maphdr}{\makebox[0.92cm][c]{mAP}}

% ---------- Combined panels ----------
\resizebox{\textwidth}{!}{%
\begin{tabular}{@{}l c|cc|cc|cc|cc|cc@{}}
\toprule
\multirow{2}{*}{Method} & \multirow{2}{*}{Venue} &
\multicolumn{2}{c|}{ELPephants \cite{korschens2019elpephants}} &
\multicolumn{2}{c|}{SealID \cite{nepovinnykh2022sealid}} &
\multicolumn{2}{c|}{GZGC (zebra) \cite{parham2017animal}} &
\multicolumn{2}{c|}{ATRW \cite{li2021atrw}} &
\multicolumn{2}{c}{GZGC (giraffe) \cite{parham2017animal}} \\
\cmidrule(lr){3-4} \cmidrule(lr){5-6} \cmidrule(lr){7-8} \cmidrule(lr){9-10} \cmidrule(lr){11-12}
 & & \rankonehdr & \maphdr & \rankonehdr & \maphdr & \rankonehdr & \maphdr & \rankonehdr & \maphdr & \rankonehdr & \maphdr \\
\midrule
Base \cite{he2021transreid} & ICCV 2021 & 31.9 & 7.9 & 79.4 & 24.8 & 12.6 & \underline{7.1} & 96.2 & 58.9 & 21.0 & 25.3 \\
TransReID \cite{he2021transreid} & ICCV 2021 & 32.3 & 8.1 & 78.6 & 23.3 & 12.2 & 6.9 & \underline{96.5} & 59.1 & 20.0 & 25.6 \\
META \cite{xu2022mimic} & ECCV 2022 & 26.4 & 5.8 & 79.0 & 20.2 & 8.0 & 3.2 & 95.7 & 51.6 & 13.5 & 9.3 \\
CLIP \cite{li2023clip} & AAAI 2023 & 28.9 & 6.8 & 77.6 & 20.8 & 11.9 & 6.9 & 95.6 & 58.1 & 22.9 & 25.7 \\
PartAware \cite{ni2023part} & ICCV 2023 & 32.0 & 7.9 & 79.7 & 24.9 & 12.4 & \underline{7.1} & 96.2 & 58.9 & 20.6 & 25.4 \\
UniReID$\dagger$ \cite{jiao2023toward} & NeurIPS 2023 & 25.2 & 6.1 & 79.4 & 23.6 & 12.0 & 6.8 & 96.1 & 55.9 & \textbf{24.7} & 26.0 \\
BAU \cite{cho2024generalizable} & NeurIPS 2024 & 13.2 & 3.7 & \underline{80.4} & \textbf{30.8} & 10.1 & 5.1 & 90.6 & 49.9 & 20.6 & 22.2 \\
AdaFreq \cite{li2024adaptive} & ECCV 2024 & \underline{33.3} & \underline{8.4} & 78.6 & 22.8 & 12.1 & 6.8 & 96.4 & 58.0 & 21.6 & 26.0 \\
ReNorm \cite{nie2024rethinking} & ECCV 2024 & 20.7 & 5.0 & 78.6 & 23.6 & 10.6 & 5.5 & 96.3 & 56.1 & \underline{23.5} & 24.1 \\
Megadescriptor$\dagger$ \cite{vcermak2024wildlifedatasets} & WACV 2024 & 32.0 & 8.2 & 76.5 & 22.1 & \underline{12.9} & 7.0 & 95.9 & 57.4 & \underline{23.5} & \underline{26.8} \\
MiewID$\dagger$ \cite{otarashvili2024miewid} & CoRR 2024 & 16.2 & 4.3 & 72.9 & 19.9 & 9.1 & 4.7 & 94.6 & 51.2 & 17.9 & 20.8 \\
CLIP-FGDI \cite{zhao2025cilp} & TIFS 2025 & 16.2 & 4.3 & 72.4 & 22.2 & 10.6 & 5.6 & 88.5 & 45.8 & 21.3 & 24.5 \\
ARBase$\dagger$ \cite{hou2024openanimals} & ICCV 2025 & 25.4 & 5.5 & 77.9 & 21.4 & 10.0 & 3.9 & 96.1 & 53.8 & 12.7 & 9.1 \\
\midrule
Ours & - & \textbf{34.4} & \textbf{9.0} & \textbf{81.5} & \underline{25.6} & \textbf{13.1} & \textbf{7.8} & \textbf{98.0} & \textbf{60.3} & 22.9 & \textbf{27.3} \\
\midrule
\multirow{2}{*}{Method} & \multirow{2}{*}{Venue} &
\multicolumn{2}{c|}{iPanda-50 \cite{wang2021giant}} &
\multicolumn{2}{c|}{HyenaID2022 \cite{wildme_hyenaid2022}} &
\multicolumn{2}{c|}{LeopardID2022 \cite{wildme_leopardid2022}} &
\multicolumn{2}{c|}{SeaTurtleID2022 \cite{Adam_2024_WACV}} &
\multicolumn{2}{c}{WhaleSharkID \cite{Holmberg_2009}} \\
\cmidrule(lr){3-4} \cmidrule(lr){5-6} \cmidrule(lr){7-8} \cmidrule(lr){9-10} \cmidrule(lr){11-12}
 & & \rankonehdr & \maphdr & \rankonehdr & \maphdr & \rankonehdr & \maphdr & \rankonehdr & \maphdr & \rankonehdr & \maphdr \\
\midrule
Base \cite{he2021transreid} & ICCV 2021 & \underline{91.8} & \underline{13.1} & 60.4 & 22.6 & 76.0 & 17.5 & 46.7 & 7.3 & 37.4 & 6.9 \\
TransReID \cite{he2021transreid} & ICCV 2021 & \underline{91.8} & \underline{13.1} & \underline{62.9} & \underline{23.7} & \underline{78.0} & \underline{18.4} & 51.2 & 8.1 & 42.2 & 7.8 \\
META \cite{xu2022mimic} & ECCV 2022 & 87.9 & 11.4 & 50.8 & 15.5 & 64.0 & 12.0 & 39.3 & 5.0 & 35.1 & 5.4 \\
CLIP \cite{li2023clip} & AAAI 2023 & 90.2 & 12.9 & 59.3 & 21.3 & 74.7 & 16.5 & 45.9 & 6.9 & 38.8 & 7.0 \\
PartAware \cite{ni2023part} & ICCV 2023 & 91.7 & \underline{13.1} & 61.8 & 23.0 & 77.6 & 18.3 & 51.1 & 8.0 & 40.2 & 7.3 \\
UniReID$\dagger$ \cite{jiao2023toward} & NeurIPS 2023 & 88.8 & 12.9 & 59.2 & 21.2 & 74.6 & 16.4 & 45.8 & 6.9 & 30.9 & 5.2 \\
BAU \cite{cho2024generalizable} & NeurIPS 2024 & 81.9 & 10.2 & 49.8 & 15.0 & 65.6 & 11.9 & 51.2 & 7.3 & 22.4 & 3.6 \\
AdaFreq \cite{li2024adaptive} & ECCV 2024 & 90.6 & 12.8 & 59.9 & 23.0 & 76.5 & 17.4 & 53.1 & 8.4 & \underline{42.5} & \underline{8.0} \\
ReNorm \cite{nie2024rethinking} & ECCV 2024 & 82.1 & 9.9 & 59.9 & 21.4 & 72.5 & 13.6 & \textbf{60.4} & \underline{9.3} & 32.9 & 5.4 \\
Megadescriptor$\dagger$ \cite{vcermak2024wildlifedatasets} & WACV 2024 & 89.6 & 13.0 & 61.3 & 23.0 & 77.1 & 18.0 & 45.3 & 6.9 & 39.1 & 7.1 \\
MiewID$\dagger$ \cite{otarashvili2024miewid} & CoRR 2024 & 74.2 & 8.9 & 53.0 & 17.1 & 66.3 & 12.0 & 42.0 & 5.5 & 27.2 & 4.5 \\
CLIP-FGDI \cite{zhao2025cilp} & TIFS 2025 & 59.9 & 6.9 & 47.8 & 15.2 & 65.8 & 13.1 & 20.2 & 3.0 & 25.9 & 4.9 \\
ARBase$\dagger$ \cite{hou2024openanimals} & ICCV 2025 & 88.2 & 11.7 & 51.1 & 15.6 & 64.3 & 12.1 & 39.6 & 5.1 & 34.4 & 5.8 \\
\midrule
Ours & - & \textbf{92.5} & \textbf{13.8} & \textbf{64.0} & \textbf{24.8} & \textbf{78.9} & \textbf{19.4} & \underline{58.8} & \textbf{10.0} & \textbf{45.4} & \textbf{8.9} \\
\bottomrule
\end{tabular}%
}
\end{table*}

\begin{figure*}[t]
  \centering
  \includegraphics[width=1.0\textwidth]{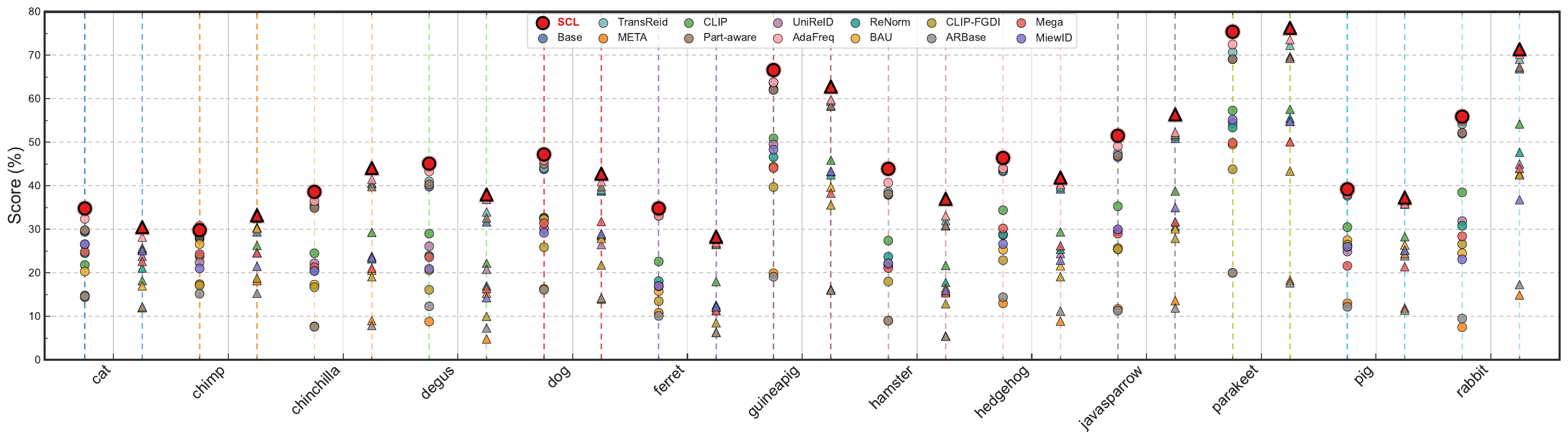}

  \caption{Protocol-1 results on PetFace (13 species). Circles: mAP; triangles: Rank-1.}
  \label{fig:petface_13species_grouped_points}

\end{figure*}

\subsection{Comparison with SOTA Methods} 

Table~\ref{tab:wildlife_complete} and Fig.~\ref{fig:petface_13species_grouped_points} report Protocol-1 results; Table~\ref{tab:wildlife71_protocol2} reports Protocol-2 results on Wildlife71. We compare four categories of methods:
(1) \textit{General ReID} (\eg, TransReID, CLIP-ReID). These methods employ strong architectures and often achieve competitive results, yet they remain species-specific and do not capture fine-grained cues that transfer across species.
(2) \textit{Animal ReID} (\eg, UniReID, AdaFreq, ARBase). These methods enhance within-species discrimination but lack cross-species structural alignment, leading to negative transfer on unseen domains. Foundation models such as MegaDescriptor \cite{vcermak2024wildlifedatasets} and MiewID \cite{otarashvili2024miewid} benefit from large-scale data but still optimize instance-level metrics without explicit cross-species alignment.
(3) \textit{DG person ReID} (\eg, ReNorm, META). These methods improve statistical invariance and handle style variations, but do not address the semantic shifts induced by species changes and therefore underperform on unknown species.
(4) \textit{DG animal ReID} (\eg, UniReID). UniReID relies on dataset-specific textual descriptions; when applied to a face-centric dataset such as PetFace, its generic whole-body description becomes mismatched, leading to a substantial performance drop.
Overall, existing methods may excel on selected species yet degrade significantly when the species changes. In contrast, our approach learns a species-agnostic model without any target-domain adaptation and achieves leading performance across diverse unseen species.

% \begin{figure*}[t]
%     \centering
%     \includegraphics[width=0.82\textwidth]{combined_hyperparameter_analysis.pdf}
%     \caption{Hyperparameter sensitivity across three panels: (a) memory size selected from \(\{1024, 2048, 4096,8192 \}\), (b) reciprocal validation size \(K_2 \in [1, 5]\), and (c) margin selected from \(\{0.1, 0.2, 0.3, 0.5\}\). The curves report mAP (\%) over six species, including seal, giraffe, tiger, chinchilla, hamster, and degus.}
%     \label{hyperparameter_analysis}
%     % \vspace{-4mm}
% \end{figure*}

\subsection{Ablation Experiments}
\textbf{Effectiveness of Each Component.} Ablation results in Table~\ref{tab:ablation_wildlife_main} confirm the complementary contributions of FDSNorm and CNM to cross-species generalization. FDSNorm stabilizes style statistics to yield domain-robust features, while CNM promotes intra-species cohesion and cross-species semantic connectivity. Table~\ref{tab:cnm_memory_ablation} further shows that CNM with memory significantly improves nearest-neighbor identity purity over vanilla k-NN; crucially, removing the memory queue leads to a noticeable drop in retrieval purity, proving it essential for providing stable, long-term relational anchors. Table~\ref{tab:layer_ema_ablation} reveals that sparse layer placement ($l \in \{0,4,8\}$) is optimal for FDSNorm---applying it too densely or across all layers over-suppresses structural semantics. Removing EMA further destabilizes feature evolution, confirming the necessity of temporal smoothing.

\begin{table*}[!t]
  \centering
  \captionsetup{font=small,skip=1pt}
  \begin{minipage}[t]{0.55\textwidth}
    \centering
    \captionof{table}{Protocol-2 results on Wildlife71.}
    \label{tab:wildlife71_protocol2}
    {\tiny
    \setlength{\tabcolsep}{5pt}
    \renewcommand{\arraystretch}{0.9}
    \resizebox{0.95\linewidth}{!}{%
    \begin{tabular}{@{}lccc@{}}
    \toprule
    Method & mAP & mINP & Rank1 \\
    \midrule
    Base \cite{he2021transreid} & 90.1 & 75.1 & 96.6 \\
    TransReID \cite{he2021transreid} & \underline{91.6} & 72.6 & 96.5 \\
    META \cite{xu2022mimic} & 83.7 & 51.0 & 96.5 \\
    CLIP \cite{li2023clip} & 86.4 & 54.1 & 96.7 \\
    PartAware \cite{ni2023part} & 90.1 & 70.3 & 96.8 \\
    UniReID$\dagger$ \cite{jiao2023toward} & 84.4 & 53.6 & 96.6 \\
    BAU \cite{cho2024generalizable} & 86.7 & 56.4 & 96.2 \\
    AdaFreq \cite{li2024adaptive} & 91.2 & \textbf{79.2} & 97.3 \\
    ReNorm \cite{nie2024rethinking} & 71.9 & 25.3 & 95.6 \\
    Megadescriptor$\dagger$ \cite{vcermak2024wildlifedatasets} & 87.3 & 60.6 & 97.3 \\
    MiewID$\dagger$ \cite{otarashvili2024miewid} & 82.5 & 43.3 & 96.1 \\
    CLIP-FGDI \cite{zhao2025cilp} & 73.6 & 27.7 & \underline{97.4} \\
    ARBase$\dagger$ \cite{hou2024openanimals} & 86.4 & 59.4 & 96.8 \\
    \midrule
    \textbf{Ours} & \textbf{93.8} & \underline{78.5} & \textbf{97.6} \\
    \bottomrule
    \end{tabular}}
    }
  \end{minipage}\hfill
  \begin{minipage}[t]{0.43\textwidth}
    \centering
    \captionof{table}{Ablation on CNM memory.}
 
    \label{tab:cnm_memory_ablation}
    {\scriptsize
    \setlength{\tabcolsep}{4pt}
    \renewcommand{\arraystretch}{1.0}
    \resizebox{\linewidth}{!}{%
    \begin{tabular}{@{}l|ccc@{}}
    \toprule
    Setting & \shortstack{Same-ID\\(\%)} & mAP (\%) & Rank1 (\%) \\
    \midrule
    k-NN & 65.12 & 18.36 & 53.53 \\
    CNM w/o Mem. & 78.24 & 19.33 & 56.01 \\
    Ours (CNM) & 89.36 & 20.69 & 58.75 \\
    \bottomrule
    \end{tabular}
    }
    }

    \captionof{table}{Layer replacement and EMA ablation results.}

    \label{tab:layer_ema_ablation}
    {
    \setlength{\tabcolsep}{6pt}
    \renewcommand{\arraystretch}{1.0}
    \resizebox{\linewidth}{!}{%
    \begin{tabular}{@{}l|cc@{}}
    \toprule
    Setting & mAP (\%) & Rank1 (\%) \\
    \midrule
    $l\in\{0,4,8\}$ & 20.69 & 58.75 \\
    $l\in\{2,4,6\}$ & 19.77 & 56.58 \\
    $l\in\{0,\ldots,11\}$ & 19.37 & 54.53 \\
    $l\in\{0,4,8\}$ w/o EMA & 20.14 & 57.53 \\
    \bottomrule
    \end{tabular}
    }
    }
  \end{minipage}
\end{table*}

\begin{table*}[t]
  \centering
  \caption{Ablation study of different components on ten wildlife datasets under Protocol-1. Intra: Intra-species consistency; Cross: Cross-species consistency.}
  \label{tab:ablation_wildlife_main}
  \makebox[\textwidth][c]{\hspace{6pt}\resizebox{\textwidth}{!}{%
    % \scriptsize
    \setlength{\tabcolsep}{3.5pt}
    \renewcommand{\arraystretch}{1.05}
    \begin{tabular}{@{}c ccc|cc|cc|cc|cc|cc@{}}
    \toprule
    \multirow{2}{*}{ID} & \multirow{2}{*}{FDSNorm} & \multirow{2}{*}{Intra} & \multirow{2}{*}{Cross} &
    \multicolumn{2}{c|}{ELPephants \cite{korschens2019elpephants}} &
    \multicolumn{2}{c|}{SealID \cite{nepovinnykh2022sealid}} &
    \multicolumn{2}{c|}{Wildlife71 \cite{jiao2023toward}} &
    \multicolumn{2}{c|}{ATRW \cite{li2021atrw}} &
    \multicolumn{2}{c}{GZGC (giraffe) \cite{parham2017animal}} \\
    \cmidrule(lr){5-6} \cmidrule(lr){7-8} \cmidrule(lr){9-10} \cmidrule(lr){11-12} \cmidrule(lr){13-14}
    & & & & Rank1 & mAP & Rank1 & mAP & Rank1 & mAP & Rank1 & mAP & Rank1 & mAP \\
    \midrule
    (a) & -- & -- & -- & 31.9 & 7.9 & 79.4 & 24.8 & 96.6 & 90.1 & 96.2 & 58.9 & 21.0 & 25.3 \\
    (b) & \checkmark & -- & -- & 32.8 & 8.4 & 80.9 & 25.3 & 97.0 & 91.4 & 97.1 & 59.3 & 22.4 & 26.8 \\
    (c) & -- & \checkmark & -- & 33.1 & 8.5 & 80.8 & 24.8 & 97.2 & 92.3 & 97.7 & 59.6 & 22.2 & 26.0 \\
    (d) & -- & \checkmark & \checkmark & 33.3 & 8.6 & 81.2 & 25.0 & 97.4 & 93.0 & 97.6 & 60.0 & 22.5 & 27.0 \\
    (e) & \checkmark & \checkmark & \checkmark & 34.2 & 9.0 & 81.3 & 25.6 & 97.6 & 93.8 & 97.8 & 60.3 & 22.7 & 27.3 \\
    \midrule
    \multicolumn{4}{@{}c|}{} &
    \multicolumn{2}{c|}{PetFace \cite{shinoda2025petface}} &
    \multicolumn{2}{c|}{HyenaID2022 \cite{wildme_hyenaid2022}} &
    \multicolumn{2}{c|}{LeopardID2022 \cite{wildme_leopardid2022}} &
    \multicolumn{2}{c|}{SeaTurtleID2022 \cite{Adam_2024_WACV}} &
    \multicolumn{2}{c}{WhaleSharkID \cite{Holmberg_2009}} \\
    \cmidrule(lr){5-6} \cmidrule(lr){7-8} \cmidrule(lr){9-10} \cmidrule(lr){11-12} \cmidrule(lr){13-14}
    (a) & -- & -- & -- & 41.9 & 39.8 & 60.4 & 22.6 & 76.0 & 17.5 & 46.7 & 7.3 & 37.4 & 6.9 \\
    (b) & \checkmark & -- & -- & 43.6 & 42.8 & 62.6 & 24.1 & 77.8 & 18.7 & 55.5 & 9.2 & 44.0 & 8.4 \\
    (c) & -- & \checkmark & -- & 44.5 & 44.8 & 63.2 & 23.9 & 78.2 & 18.6 & 56.7 & 8.9 & 44.3 & 8.5 \\
    (d) & -- & \checkmark & \checkmark & 45.3 & 45.9 & 63.5 & 24.2 & 78.5 & 18.9 & 57.4 & 9.3 & 44.7 & 8.7 \\
    (e) & \checkmark & \checkmark & \checkmark & 46.2 & 46.9 & 63.8 & 24.8 & 78.7 & 19.4 & 58.6 & 10.0 & 45.2 & 8.9 \\
    \bottomrule
    \end{tabular}%
  }}

\end{table*}

\noindent\textbf{Foreground-Only Analysis}. We use MVANet \cite{yu2024mvanet} to segment foreground regions and construct foreground-only inputs. Table~\ref{tab:wildlife71_foreground_background_compare} further evaluates robustness when background cues are largely removed. All methods degrade under foreground-only input, indicating that context still contributes to matching. Nevertheless, our method achieves the strongest absolute foreground performance and the smallest mAP drop, showing that SCL relies more on transferable identity structure than on scene-specific shortcuts. 
% \textbf{Hyperparameters Analysis}. Unless specified, we use memory size \(= 4096\), \(K_1 = 8\), \(K_2 = 3\), \(\lambda_{\text{intra}} = 1\), \(\lambda_{\text{cross}} = 1\), and margin \(= 0.2\). Our hyperparameter sweeps show: (1) Increasing memory size initially improves performance by providing richer candidate sets. (2) The model performs best when \(K_2 \in \{3, 4\}\), indicating a balanced reciprocal filtering strength. A larger \(K_2\) risks including false neighbors, causing negative transfer. (3) Regarding the margin, small values (\eg, 0.1) are too permissive, limiting hard-case robustness. Large values (\eg, \(\ge 0.4\)) are too stringent, hurting recall. A margin of \(0.2\) delivers the best trade-off, filtering spurious matches while preserving transferable semantics.

\begin{table*}[t]
  \centering
  \scriptsize
  \caption{Original vs. foreground-only input under Protocol-2.}

\label{tab:wildlife71_foreground_background_compare}
  \setlength{\tabcolsep}{4pt}
  \renewcommand{\arraystretch}{1.02}
  \begin{tabular}{@{}l|ccc|ccc|ccc@{}}
  \toprule
  \multirow{2}{*}{Model} &
  \multicolumn{3}{c|}{Wildlife71 \cite{jiao2023toward}} &
  \multicolumn{3}{c|}{Wildlife71 (Foreground) \cite{jiao2023toward}} &
  \multicolumn{3}{c}{Drop} \\
  \cmidrule(lr){2-4} \cmidrule(lr){5-7} \cmidrule(lr){8-10}
   & mAP & mINP & Rank1 & mAP & mINP & Rank1 & mAP & mINP & Rank1 \\
  \midrule
  Base \cite{he2021transreid} & 90.1 & 75.1 & 96.6 & 71.2 & 28.6 & 95.8 & 18.9 & 46.5 & 0.8 \\
  Megadescriptor \cite{vcermak2024wildlifedatasets} & 87.3 & 60.6 & 97.3 & 69.8 & 28.7 & 95.2 & 17.5 & 31.9 & 2.1 \\
  SCL (Ours) & 93.8 & 78.5 & 97.6 & 79.0 & 45.8 & 96.1 & 14.8 & 32.7 & 1.5 \\
  \bottomrule
  \end{tabular}

\end{table*}

\begin{figure*}[!ht]
  \centering
  \setlength{\tabcolsep}{2pt}
  \newcommand{\vispanel}[1]{\includegraphics[width=0.234\textwidth,height=1.325cm]{#1}}
  \begin{tabular}{cccc}
    \vispanel{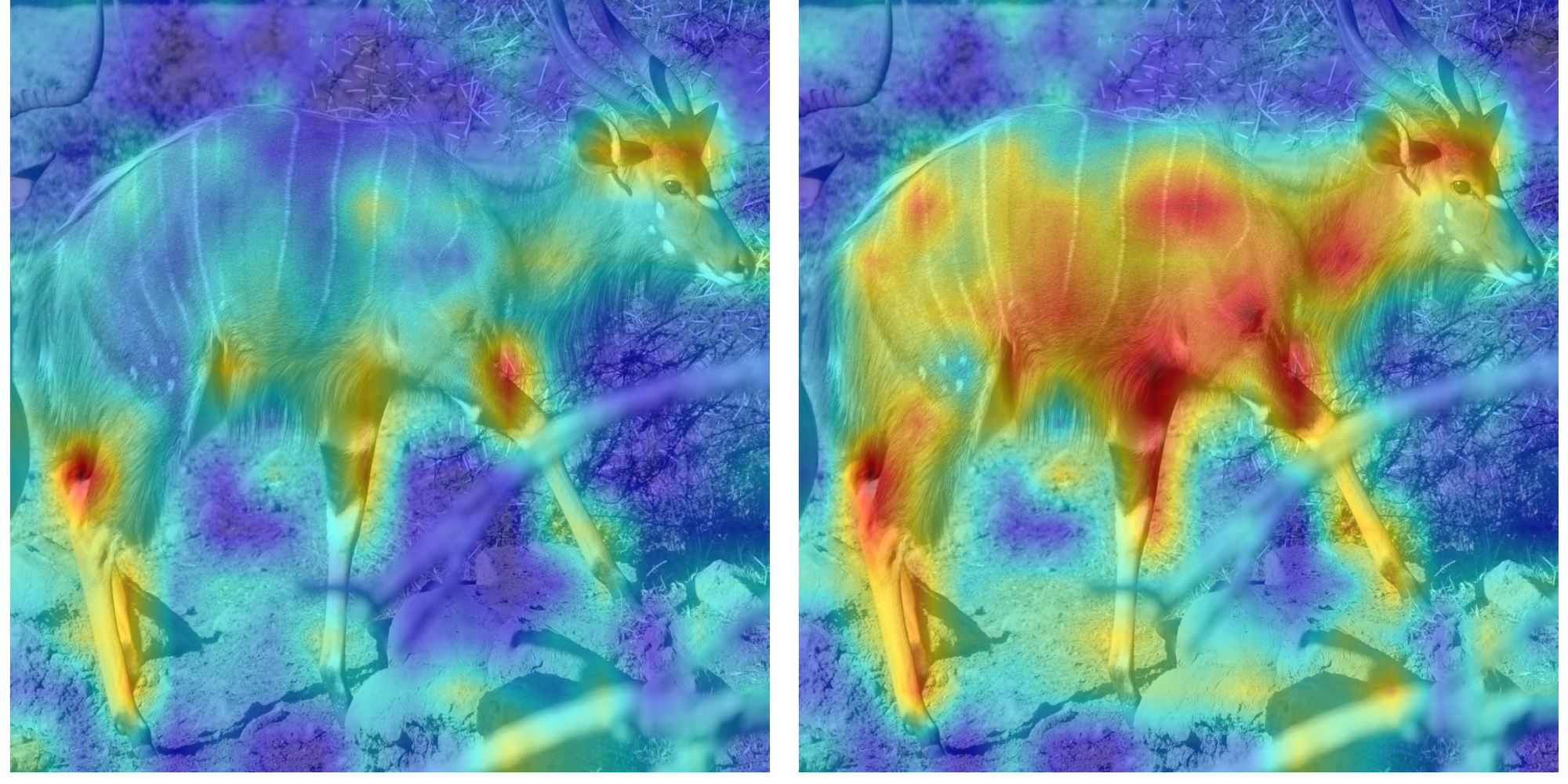} &
    \vispanel{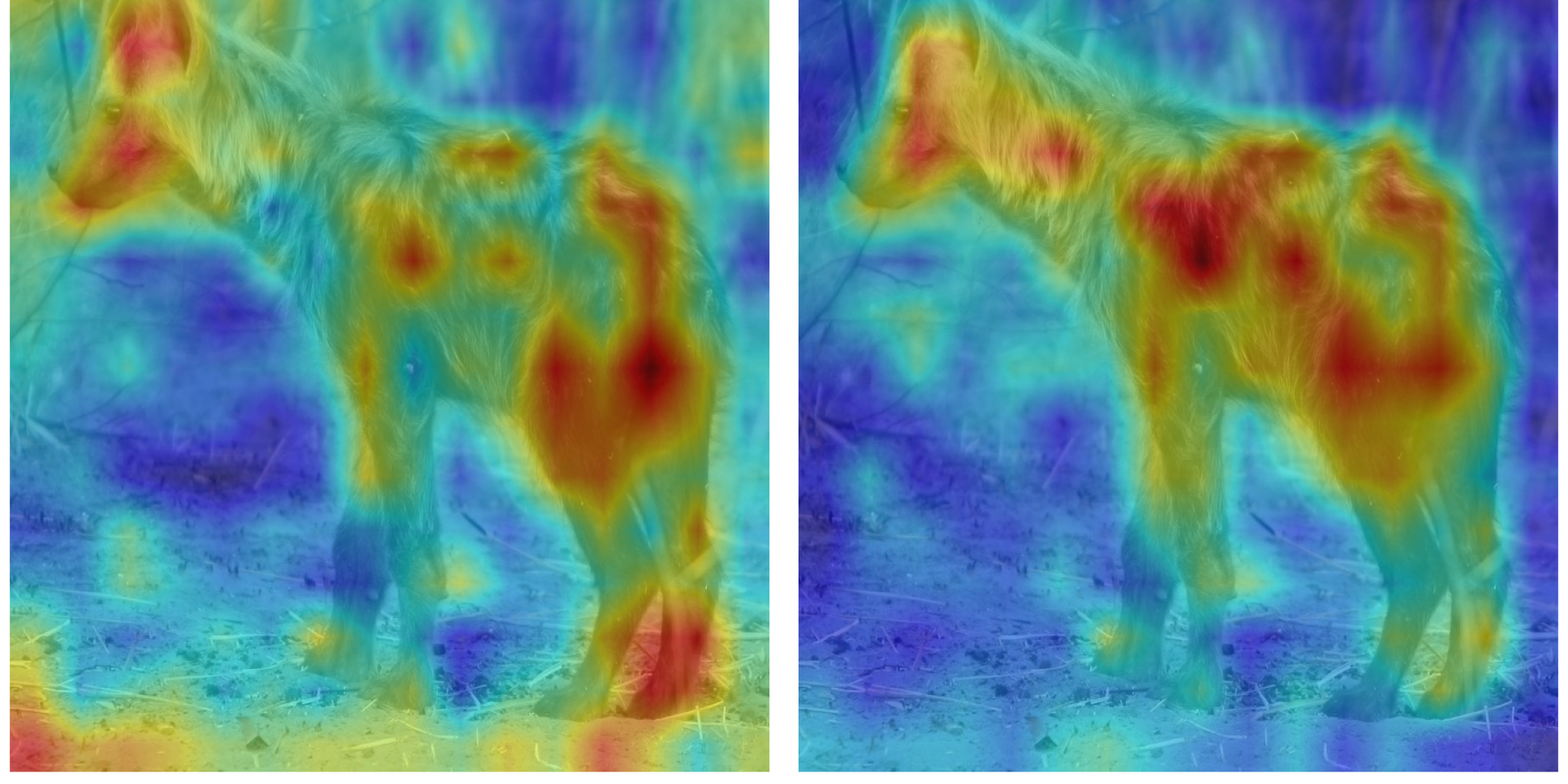} &
    \vispanel{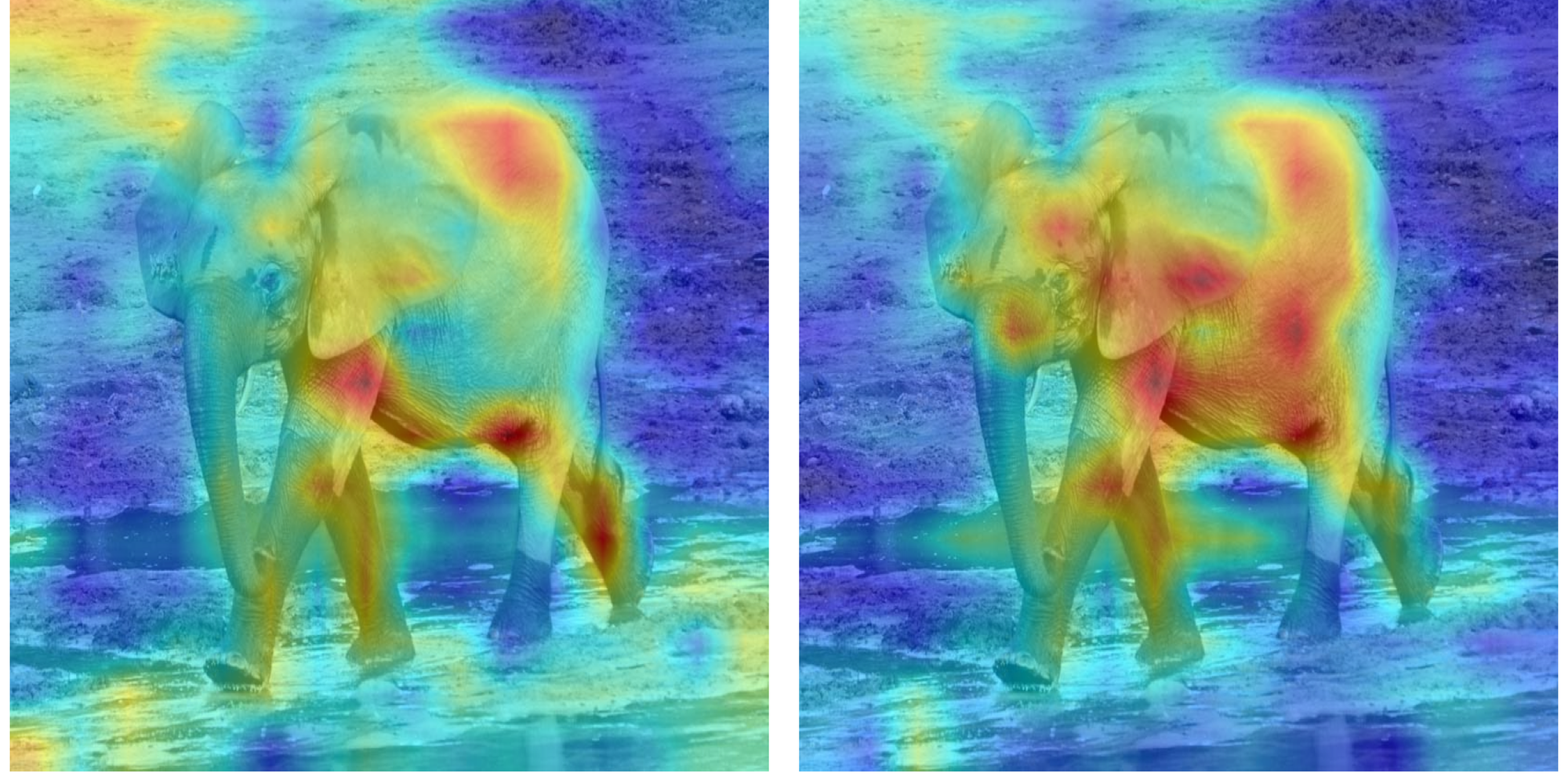} &
    \vispanel{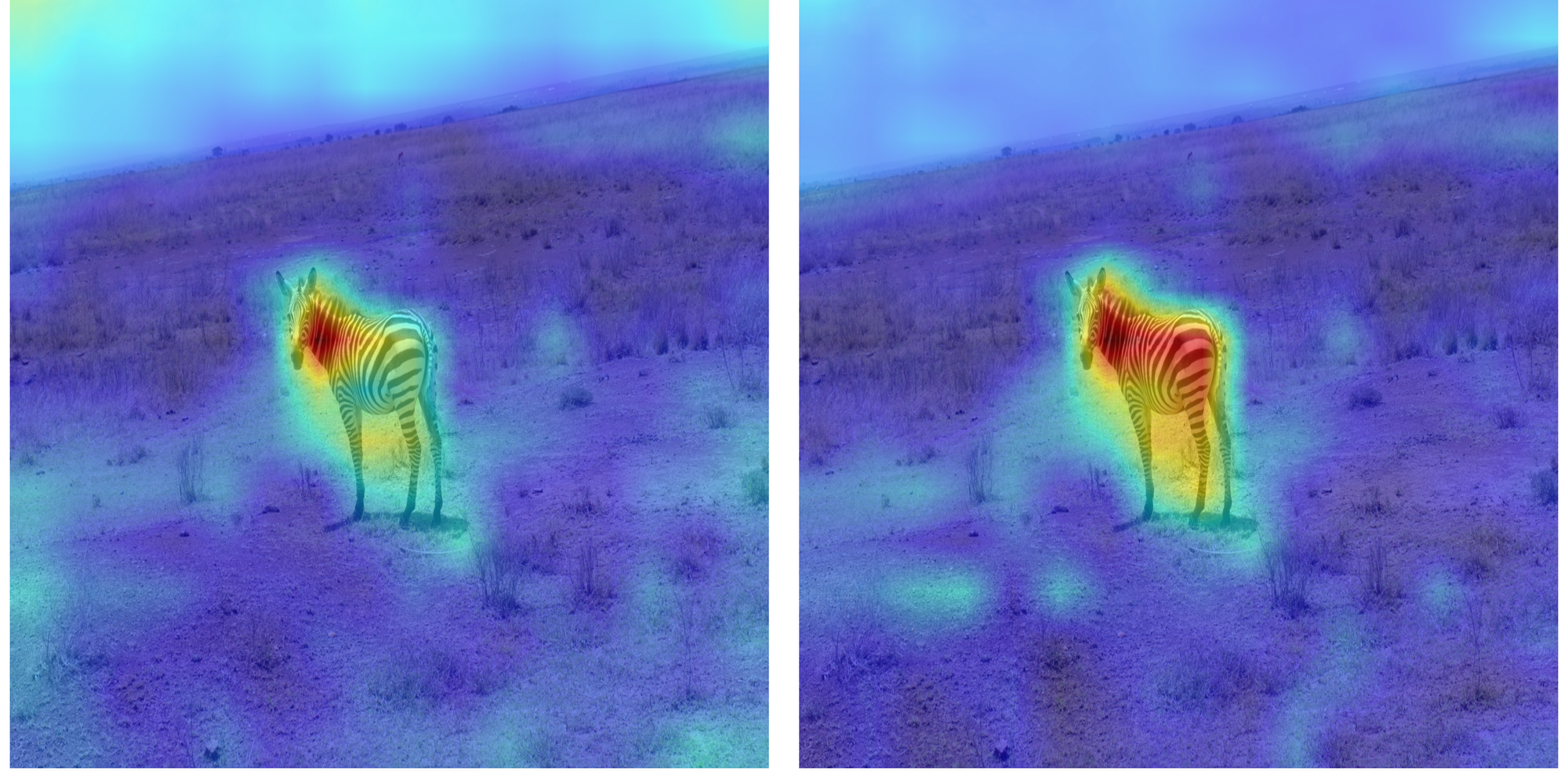} \\
    \tiny (a) Nyala & \tiny (b) Hyena & \tiny (c) Elephant & \tiny (d) Zebra
  \end{tabular}
  \begin{tabular}{cccc}
    \vispanel{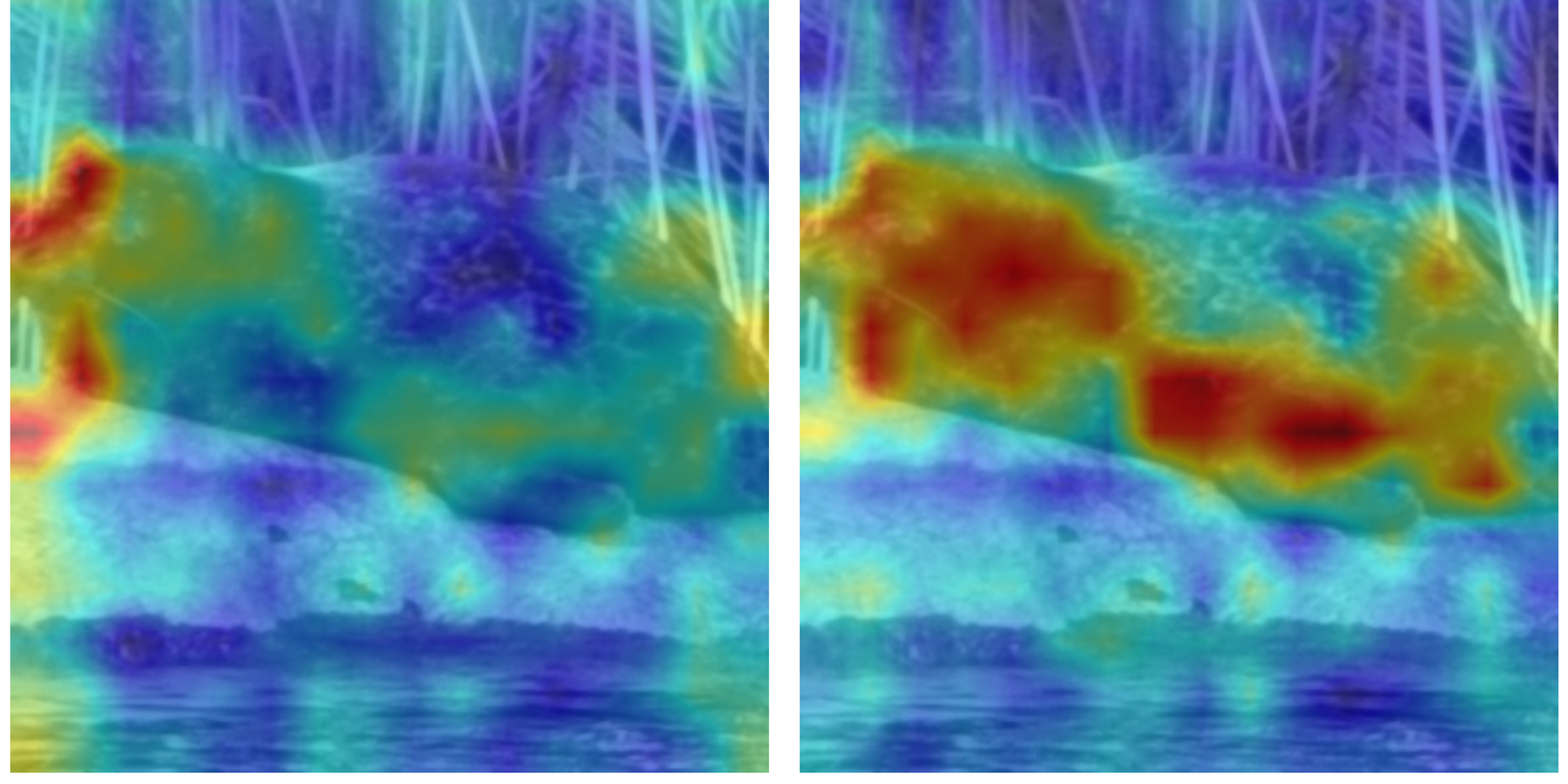} &
    \vispanel{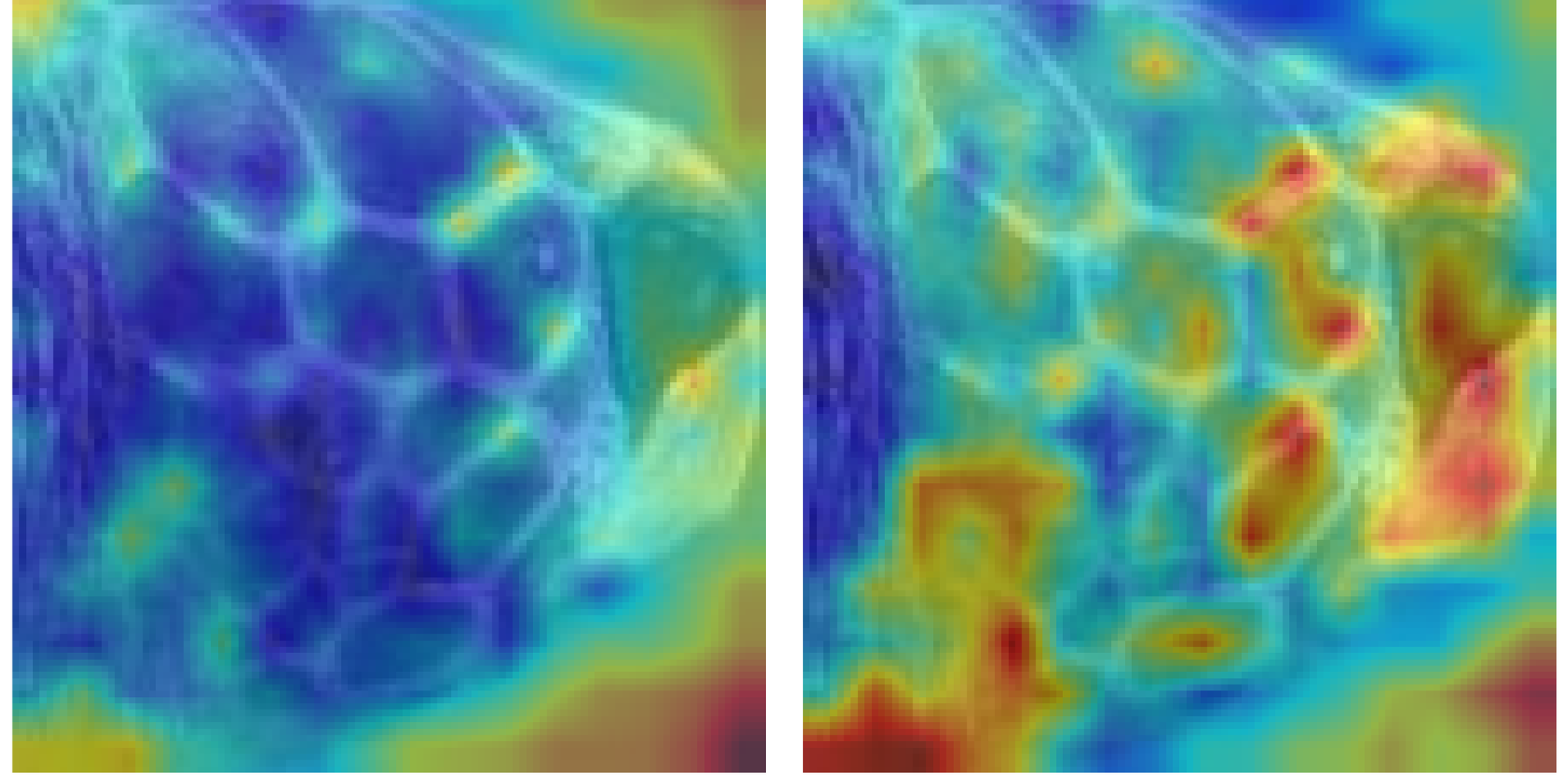} &
    \vispanel{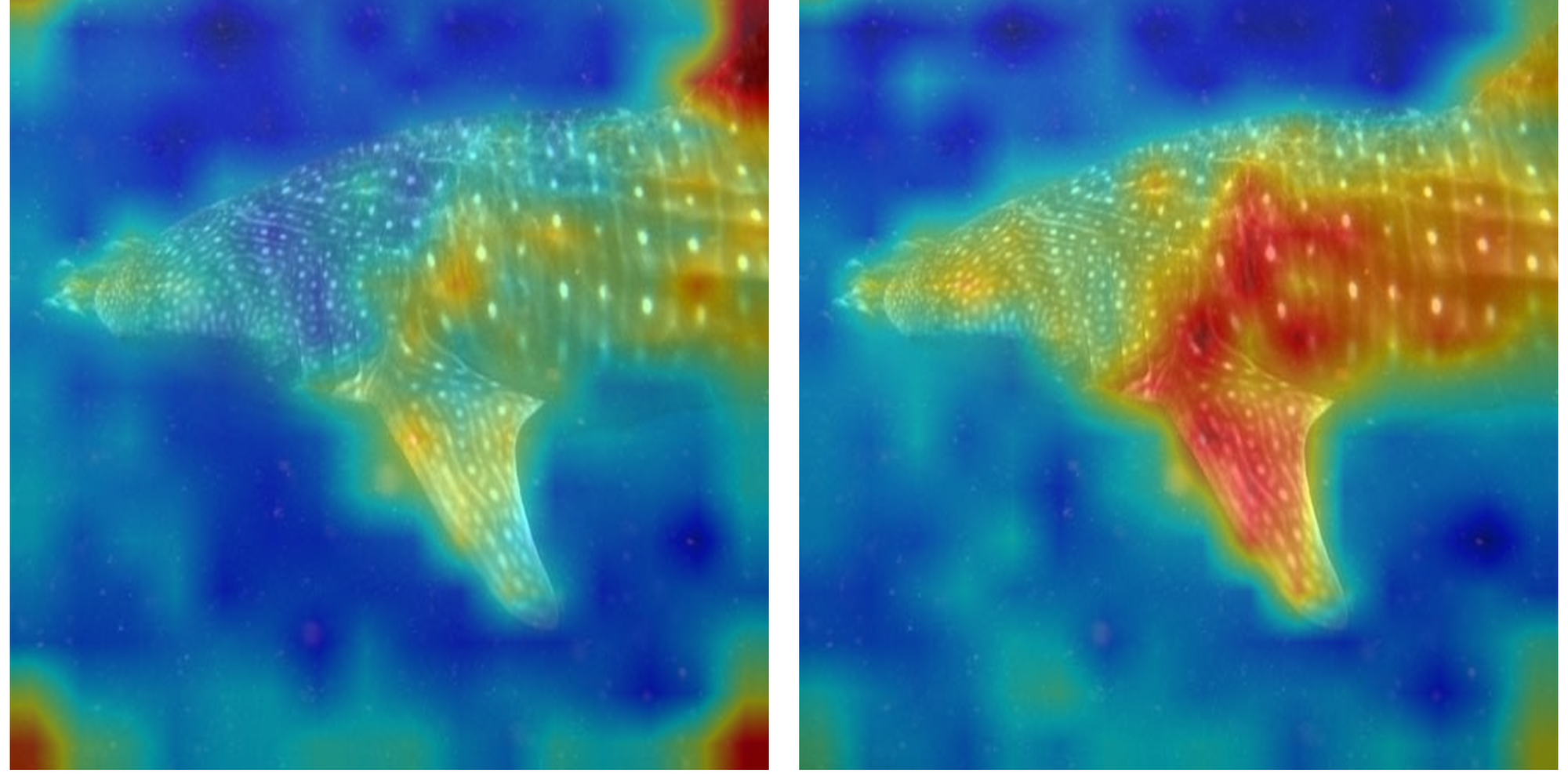} &
    \vispanel{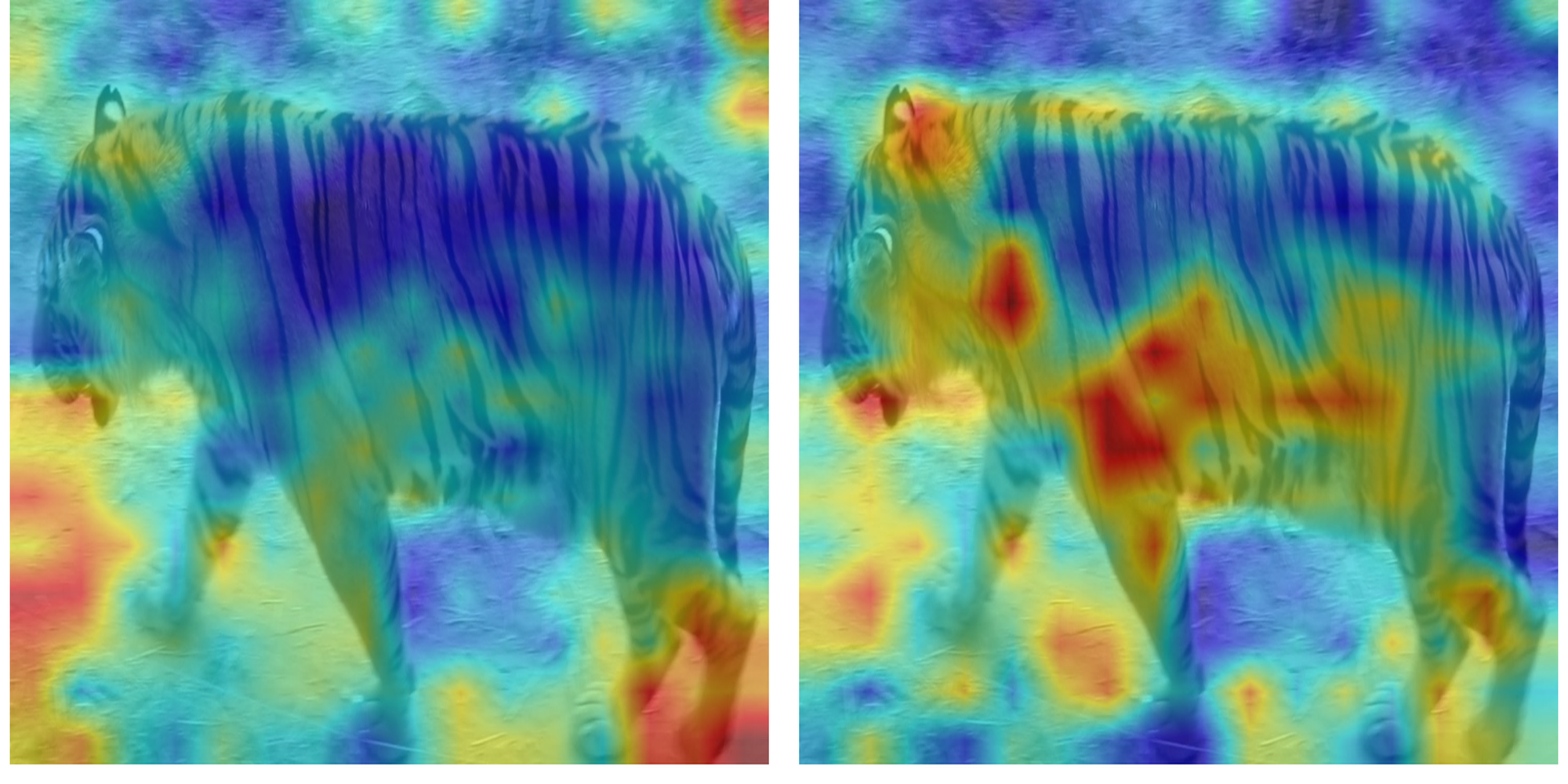} \\
    \tiny (e) Seal & \tiny (f) SeaTurtle & \tiny (g) Whale Shark & \tiny (h) Tiger
  \end{tabular}
  \caption{Last layer activation-map comparison between Base (left) and SCL (right).}
  \label{visualization}
\end{figure*}

\noindent\textbf{Visualization Analysis.}
\Cref{visualization} shows eight subfigures: Compared with Base, SCL consistently shifts attention toward semantically meaningful animal regions (\eg, torso contours, texture-rich parts, and limbs) and reduces diffuse activation on irrelevant background areas. This trend is stable across elephant, zebra, seal, sea turtle, whale shark, hyena, nyala and tiger, indicating stronger structure-focused consistency under large appearance and habitat variations.

\begin{figure*}[t]
\centering

%================ Left: Figure =================%
\begin{minipage}[t]{0.53\textwidth}
\vspace{0pt}
\centering

\includegraphics[width=\linewidth]{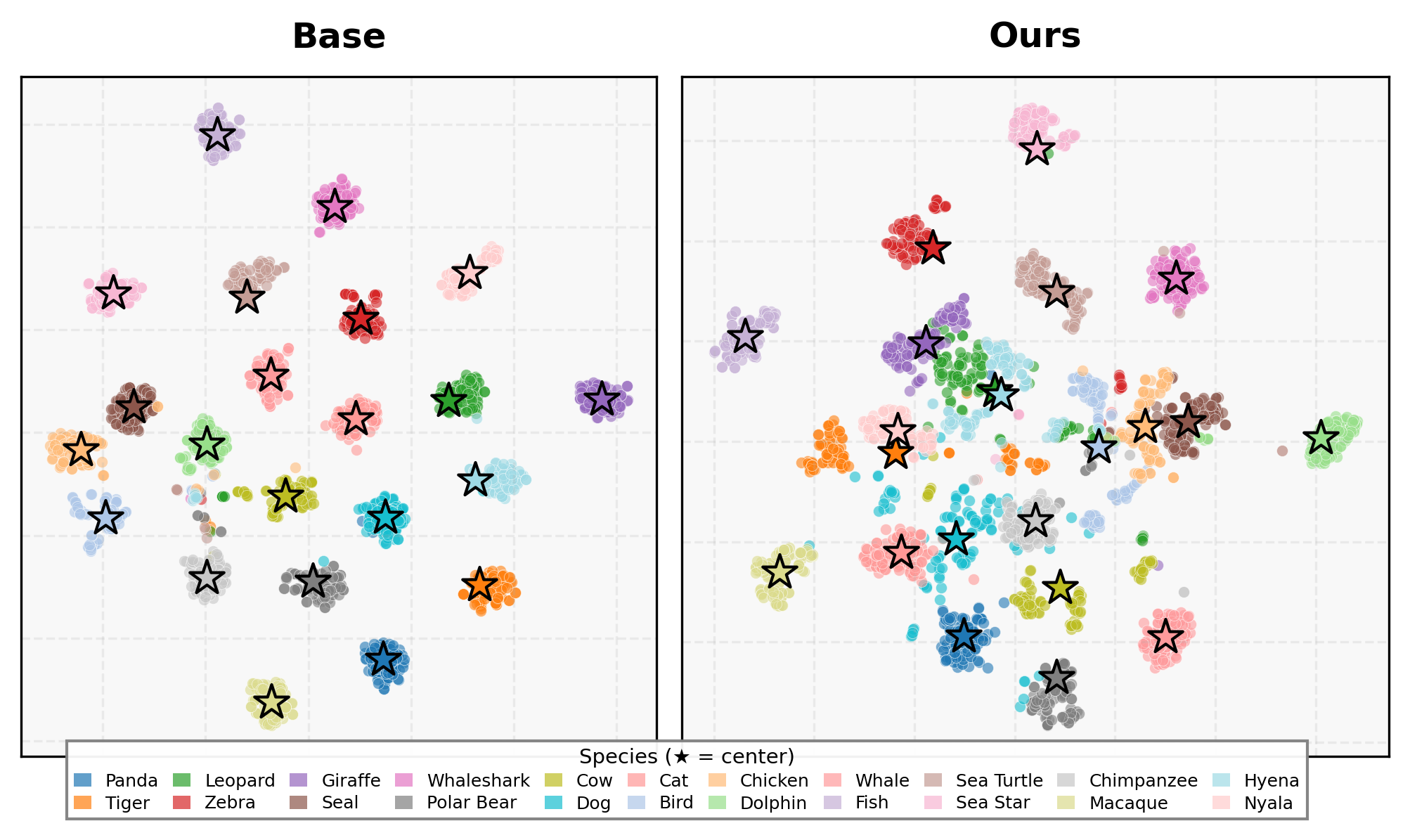}

\caption{Embedding space visualization of samples from diverse species.}
\label{fig:diverse_species}

\end{minipage}
\hfill
%================ Right: Table =================%
\begin{minipage}[t]{0.43\textwidth}
\vspace{0pt}
\centering

\captionsetup{type=table}
\caption{Ratio of inter-ID distance to intra-ID distances on Protocol-1. Higher is better.}
\label{tab:feature_distribution}

\vspace{3pt}

\scriptsize
\setlength{\tabcolsep}{3pt}

\resizebox{\linewidth}{!}{%
\begin{tabular}{lccc}
\toprule
Dataset & SCL & Mega & MiewID \\
\midrule
iPanda-50 \cite{wang2021giant} & 1.265 & 1.145 & 1.062 \\
ELPephants \cite{korschens2019elpephants} & 1.220 & 1.122 & 1.049 \\
SealID \cite{nepovinnykh2022sealid} & 1.379 & 1.263 & 1.134 \\
GZGC-Zebra \cite{parham2017animal} & 1.242 & 1.106 & 1.048 \\
ATRW \cite{li2021atrw} & 2.816 & 2.106 & 1.269 \\
GZGC-Giraffe \cite{parham2017animal} & 1.838 & 1.726 & 1.326 \\
HyenaID2022 \cite{wildme_hyenaid2022} & 1.741 & 1.412 & 1.126 \\
LeopardID2022 \cite{wildme_leopardid2022} & 1.456 & 1.272 & 1.091 \\
SeaTurtleID2022 \cite{Adam_2024_WACV} & 1.311 & 1.234 & 1.108 \\
WhaleSharkID \cite{Holmberg_2009} & 1.196 & 1.064 & 1.039 \\
\bottomrule
\end{tabular}
}

\end{minipage}

\end{figure*}

\noindent\textbf{Feature Distribution Analysis.}
Figure~\ref{fig:diverse_species} shows what CNM learns in the embedding space. For this experiment, Base denotes the plain ViT baseline. Both the baseline and our model are trained on Wildlife71, and we randomly sample 100 instances for each of the 22 species for visualization.  The Base representation exhibits clear species-wise separation, where samples from different species form isolated clusters. In contrast, CNM produces a more mixed embedding distribution across species while achieving better ReID performance, suggesting that it reduces species-specific clustering tendencies and learns more effective cross-species representations.

% \begin{figure}[t]
%   \centering
%   \includegraphics[width=1.0\textwidth]{img/old.png}
%   \caption{Protocol-1 results on PetFace (13 species).}
%   \label{fig:supp_visualization}
% \end{figure}

% \begin{table}[!ht]
% \centering
% \caption{Ratio of inter-ID distance to intra-ID distances on Protocol-1. Higher is better.}
% \label{tab:supp_feature_distribution}
% \scriptsize
% \setlength{\tabcolsep}{5pt}
% \resizebox{0.80\linewidth}{!}{%
% \begin{tabular}{@{}lccc@{}}
% \toprule
% Dataset & SCL(Ours) & Megadescriptor & MiewID \\
% \midrule
% iPanda-50 \cite{wang2021giant} & 1.265 & 1.145 & 1.062 \\
% ELPephants \cite{korschens2019elpephants} & 1.220 & 1.122 & 1.049 \\
% SealID \cite{nepovinnykh2022sealid} & 1.379 & 1.263 & 1.134 \\
% GZGC-Zebra \cite{parham2017animal} & 1.242 & 1.106 & 1.048 \\
% ATRW \cite{li2021atrw} & 2.816 & 2.106 & 1.269 \\
% GZGC-Giraffe \cite{parham2017animal} & 1.838 & 1.726 & 1.326 \\
% HyenaID2022 \cite{wildme_hyenaid2022} & 1.741 & 1.412 & 1.126 \\
% LeopardID2022 \cite{wildme_leopardid2022} & 1.456 & 1.272 & 1.091 \\
% SeaTurtleID2022 \cite{Adam_2024_WACV} & 1.311 & 1.234 & 1.108 \\
% WhaleSharkID \cite{Holmberg_2009} & 1.196 & 1.064 & 1.039 \\
% \bottomrule
% \end{tabular}%
% }
% \end{table}

To further examine representation quality under unseen domains, we report the ratio of inter-ID distance to intra-ID distance in Table~\ref{tab:feature_distribution}. A higher ratio indicates a better clustering structure, namely tighter intra-identity compactness together with clearer inter-identity separation. Compared with strong animal foundation models such as Megadescriptor and MiewID, SCL consistently achieves higher ratios on all ten unseen datasets. This result supports the same conclusion as our qualitative visualizations: our method improves cluster compactness while preserving clearer boundaries between identities, thereby mitigating species-isolated feature fragmentation.

\section{Conclusion}
This paper introduced the Semantic Consistency Learning framework for cross-species generalization in animal ReID. Through Foreground--Background Decoupled Spectral Normalization, SCL suppresses feature instability caused by environmental variations while aggregating long-range semantic cues shared across species. Cross-species Neighborhood Modeling further shifts the objective from instance-level discrimination to relational semantic understanding, enabling the model to capture structural regularities that generalize across species and yield a unified, transferable embedding space.
Promising future directions include extending SCL to fully open-set ReID, integrating large-scale ecological foundation models, and enabling adaptive deployment across diverse platforms such as UAVs and camera traps. We hope this work encourages the community to move beyond species-specific ReID and toward general, cross-species visual understanding that can support large-scale biodiversity monitoring in real-world ecosystems.

% This paper introduced the Semantic Consistency Learning framework for cross-species generalization in animal ReID. Through Foreground--Background Decoupled Spectral Normalization, SCL suppresses feature instability caused by environmental variations while aggregating long-range semantic cues shared across species. Cross-species Neighborhood Modeling further shifts the objective from instance-level discrimination to relational semantic understanding, enabling the model to capture structural regularities that generalize across species and yield a unified, transferable embedding space.
% Promising future directions include extending SCL to fully open-set ReID, integrating large-scale ecological foundation models, and enabling adaptive deployment across diverse platforms such as UAVs and camera traps. We hope this work encourages the community to move beyond species-specific ReID and toward general, cross-species visual understanding that can support large-scale biodiversity monitoring in real-world ecosystems.

% \clearpage

\bigskip

\noindent\textbf{Acknowledgments.}
This work was partially supported by the National Natural Science Foundation of China under Grant T2541022.

{ 
    \small
    \bibliographystyle{splncs04} % ECCV uses splncs04 style
    \bibliography{main}

\begin{thebibliography}{10}
\providecommand{\url}[1]{\texttt{#1}}
\providecommand{\urlprefix}{URL }
\providecommand{\doi}[1]{https://doi.org/#1}

\bibitem{Adam_2024_WACV}
Adam, L., {\v{C}}erm{\'a}k, V., Papafitsoros, K., Picek, L.: Seaturtleid2022: A
  long-span dataset for reliable sea turtle re-identification. In: WACV. pp.
  7146--7156 (2024)

\bibitem{adam2025animalclef}
Adam, L., Papafitsoros, K., Kov{\'a}{\v{r}}, R., {\v{C}}erm{\'a}k, V., Picek,
  L.: Overview of {AnimalCLEF} 2025: Recognizing individual animals in images.
  In: Working Notes of {CLEF} (2025)

\bibitem{andrew2021friesian}
Andrew, W., Hannuna, S., Campbell, N., Burghardt, T.: Friesian: A novel dataset
  and a two-stage deep learning framework for cattle re-identification. In:
  ICIP. pp. 3103--3107 (2021)

\bibitem{Bai2021DMGNet}
Bai, Y., Jiao, J., Ce, W., Liu, J., Lou, Y., Feng, X., Duan, L.Y.: Person30k: A
  dual-meta generalization network for person re-identification. In: CVPR
  (2021)

\bibitem{wildme_hyenaid2022}
{Botswana Predator Conservation Trust}: Panthera pardus csv custom export
  (2022), \url{https://lila.science/datasets/hyena-id-2022}, retrieved from
  African Carnivore Wildbook. Dataset export dated 2022-04-28. Accessed: June
  29, 2026

\bibitem{wildme_leopardid2022}
{Botswana Predator Conservation Trust}: Panthera pardus csv custom export
  (2022), \url{https://lila.science/datasets/leopard-id-2022}, retrieved from
  African Carnivore Wildbook. Dataset export dated 2022-04-28. Accessed: June
  29, 2026

\bibitem{vcermak2024wildlifedatasets}
{\v{C}}erm{\'a}k, V., Picek, L., Adam, L., Papafitsoros, K.: Wildlifedatasets:
  An open-source toolkit for animal re-identification. In: WACV. pp. 5953--5963
  (2024)

\bibitem{chen2026object}
Chen, S., Wu, Y., Ye, M.: Object-generalized re-identification: A step towards
  universal instance perception. In: CVPR. pp. 18481--18491 (2026)

\bibitem{chen2022rotation}
Chen, S., Ye, M., Du, B.: Rotation invariant transformer for recognizing object
  in uavs. In: ACM MM. pp. 2565--2574 (2022)

\bibitem{chen2017person}
Chen, Y.C., Zhu, X., Zheng, W.S., Lai, J.H.: Person re-identification by camera
  correlation aware feature augmentation. IEEE TPAMI  \textbf{40}(2),  392--408
  (2017)

\bibitem{cho2024generalizable}
Cho, Y., Kim, J., Kim, W.J., Jung, J., eui Yoon, S.: Generalizable person
  re-identification via balancing alignment and uniformity. In: NeurIPS (2024)

\bibitem{choi2021meta}
Choi, S., Kim, T., Jeong, M., Park, H., Kim, C.: Meta batch-instance
  normalization for generalizable person re-identification. In: CVPR. pp.
  3425--3435 (2021)

\bibitem{Dai2021RaMoE}
Dai, Y., Li, X., Liu, J., Tong, Z., Duan, L.Y.: Generalizable person
  re-identification with relevance-aware mixture of experts. In: CVPR. pp.
  16145--16154 (2021)

\bibitem{dosovitskiy2020image}
Dosovitskiy, A., Beyer, L., Kolesnikov, A., Weissenborn, D., Zhai, X.,
  Unterthiner, T., Dehghani, M., Minderer, M., Heigold, G., Gelly, S., et~al.:
  An image is worth 16x16 words: Transformers for image recognition at scale.
  ICLR  (2020)

\bibitem{he2021transreid}
He, S., Luo, H., Wang, P., Wang, F., Li, H., Jiang, W.: Transreid:
  Transformer-based object re-identification. In: ICCV. pp. 15013--15022 (2021)

\bibitem{hermans2017defense}
Hermans, A., Beyer, L., Leibe, B.: In defense of the triplet loss for person
  re-identification. arXiv preprint arXiv:1703.07737  (2017)

\bibitem{Holmberg_2009}
Holmberg, J., Norman, B., Arzoumanian, Z.: Estimating population size,
  structure, and residency time for whale sharks rhincodon typus through
  collaborative photo-identification. Endangered Species Research
  \textbf{7}(1),  39--53 (2009)

\bibitem{hou2024openanimals}
Hou, S., Huang, P., Wang, Z., Liu, Y., Li, Z., Zhang, M., Huang, Y.:
  Openanimals: Revisiting person re-identification for animals towards better
  generalization. ICCV  (2024)

\bibitem{jiang2024domain}
Jiang, Y., Cheng, X., Yu, H., Liu, X., Chen, H., Zhao, G.: Domain shifting: A
  generalized solution for heterogeneous cross-modality person
  re-identification. In: ECCV. pp. 289--306 (2024)

\bibitem{jiao2023toward}
Jiao, B., Liu, L., Gao, L., Wu, R., Lin, G., Wang, P., Zhang, Y.: Toward
  re-identifying any animal. NeurIPS  \textbf{36},  40042--40053 (2023)

\bibitem{jin2020style}
Jin, X., Lan, C., Zeng, W., Chen, Z., Zhang, L.: Style normalization and
  restitution for generalizable person re-identification. In: CVPR. pp.
  3143--3152 (2020)

\bibitem{korschens2019elpephants}
Korschens, M., Denzler, J.: Elpephants: A fine-grained dataset for elephant
  re-identification. In: ICCVW. pp.~0--0 (2019)

\bibitem{lee2025domain}
Lee, H., Park, J., Oh, J., Eom, C.: Domain generalization for person
  re-identification: A survey towards domain-agnostic person matching.
  Neurocomputing p. 130763 (2025)

\bibitem{lee2023decompose}
Lee, S., Bae, J., Kim, H.Y.: Decompose, adjust, compose: Effective
  normalization by playing with frequency for domain generalization. In: CVPR.
  pp. 11776--11785 (2023)

\bibitem{li2024adaptive}
Li, C., Chen, S., Ye, M.: Adaptive high-frequency transformer for diverse
  wildlife re-identification. In: ECCV. pp. 296--313. Springer (2024)

\bibitem{li2021atrw}
Li, S., Li, J.W., Wu, C., Zheng, W.S.: {ATRW}: A benchmark for amur tiger
  re-identification in the wild. In: ACM MM. pp. 1297--1305 (2021)

\bibitem{li2023clip}
Li, S., Sun, L., Li, Q.: Clip-reid: exploiting vision-language model for image
  re-identification without concrete text labels. In: AAAI. vol.~37, pp.
  1405--1413 (2023)

\bibitem{Liao2020QAConv}
Liao, S., Shao, L.: Interpretable and generalizable person re-identification
  with query-adaptive convolution and temporal lifting. In: ECCV. pp. 456--474.
  Springer (2020)

\bibitem{lin2021domain}
Lin, C., Yuan, Z., Zhao, S., Sun, P., Wang, C., Cai, J.: Domain-invariant
  disentangled network for generalizable object detection. In: ICCV. pp.
  8771--8780 (2021)

\bibitem{lin2023deep}
Lin, S., Zhang, Z., Huang, Z., Lu, Y., Lan, C., Chu, P., You, Q., Wang, J.,
  Liu, Z., Parulkar, A., et~al.: Deep frequency filtering for domain
  generalization. In: CVPR. pp. 11797--11807 (2023)

\bibitem{lou2019veri}
Lou, Y., Bai, Y., Liu, J., Wang, S., Duan, L.: Veri-wild: A large dataset and a
  new method for vehicle re-identification in the wild. In: CVPR. pp.
  3235--3243 (2019)

\bibitem{luo2019bag}
Luo, H., Gu, Y., Liao, X., Lai, S., Jiang, W.: Bag of tricks and a strong
  baseline for deep person re-identification. In: CVPRW. pp. 122--130 (2019)

\bibitem{nepovinnykh2024species}
Nepovinnykh, E., Chelak, I., Eerola, T., Immonen, V., K{\"a}lvi{\"a}inen, H.,
  Kholiavchenko, M., Stewart, C.V.: Species-agnostic patterned animal
  re-identification by aggregating deep local features. IJCV  \textbf{132}(9),
  4003--4018 (2024)

\bibitem{nepovinnykh2022sealid}
Nepovinnykh, E., Eerola, T., Biard, V., Mutka, P., Niemi, M., Kunnasranta, M.,
  K{\"a}lvi{\"a}inen, H.: Sealid: Saimaa ringed seal re-identification dataset.
  Sensors  \textbf{22}(19), ~7602 (2022)

\bibitem{nguyen2024tackling}
Nguyen, V.D., Mirza, S., Zakeri, A., Gupta, A., Khaldi, K., Aloui, R., Mantini,
  P., Shah, S.K., Merchant, F.: Tackling domain shifts in person
  re-identification: A survey and analysis. In: CVPRW. pp. 4149--4159 (2024)

\bibitem{ni2023part}
Ni, H., Li, Y., Gao, L., Shen, H.T., Song, J.: Part-aware transformer for
  generalizable person re-identification. In: ICCV. pp. 11280--11289 (2023)

\bibitem{ni2022meta}
Ni, H., Song, J., Luo, X., Zheng, F., Li, W., Shen, H.T.: Meta distribution
  alignment for generalizable person re-identification. In: CVPR. pp.
  2487--2496 (2022)

\bibitem{nie2024rethinking}
Nie, R., Ding, J., Zhou, X., Li, X.: Rethinking normalization layers for domain
  generalizable person re-identification. In: ECCV. pp. 267--284. Springer
  (2024)

\bibitem{otarashvili2024miewid}
Otarashvili, L., Subramanian, T., Holmberg, J., Levenson, J.J., Stewart, C.V.:
  Multispecies animal re-id using a large community-curated dataset. CoRR
  (2024)

\bibitem{papafitsoros2022seaturtleid}
Papafitsoros, K., Adam, L., {\v{C}}erm{\'a}k, V., Picek, L.: Seaturtleid: A
  novel long-span dataset highlighting the importance of timestamps in wildlife
  re-identification. arXiv preprint arXiv:2211.10307  (2022)

\bibitem{parham2017animal}
Parham, J., Crall, J., Stewart, C., Berger-Wolf, T., Rubenstein, D.I.: Animal
  population censusing at scale with citizen science and photographic
  identification. In: AAAI (2017)

\bibitem{shinoda2025petface}
Shinoda, R., Shiohara, K.: Petface: A large-scale dataset and benchmark for
  animal identification. In: ECCV. pp. 19--36. Springer (2025)

\bibitem{Song2019DIMN}
Song, J., Yang, Y., Li, Y.Z., Hospedales, T.M.: Generalizable person
  re-identification by domain-invariant mapping network. In: CVPR. pp. 718--727
  (2019)

\bibitem{wang2018learning}
Wang, G., Yuan, Y., Chen, X., Li, J., Zhou, X.: Learning discriminative
  features with multiple granularities for person re-identification. In: ACM
  MM. pp. 274--282 (2018)

\bibitem{wang2021giant}
Wang, L., Ding, R., Zhai, Y., Zhang, Q., Tang, W., Zheng, N., Hua, G.: Giant
  panda identification. IEEE TIP  \textbf{30},  2837--2849 (2021)

\bibitem{xu2022mimic}
Xu, B., Liang, J., He, L., Sun, Z.: Mimic embedding via adaptive aggregation:
  Learning generalizable person re-identification. In: ECCV. pp. 372--388.
  Springer (2022)

\bibitem{yang2024pedestrian}
Yang, Z., Wu, D., Wu, C., Lin, Z., Gu, J., Wang, W.: A pedestrian is worth one
  prompt: Towards language guidance person re-identification. In: CVPR. pp.
  17343--17353 (2024)

\bibitem{ye2024transformer}
Ye, M., Chen, S., Li, C., Zheng, W.S., Crandall, D., Du, B.: Transformer for
  object re-identification: A survey. arXiv preprint arXiv:2401.06960  (2024)

\bibitem{yu2024mvanet}
Yu, Q., Zhao, X., Pang, Y., Zhang, L., Lu, H.: Multi-view aggregation network
  for dichotomous image segmentation. In: CVPR. pp. 3921--3930 (2024)

\bibitem{yuan2025from}
Yuan, C., Zhang, G., Ma, C., Zhang, T., Niu, G.: From poses to identity:
  Training-free person re-identification via feature centralization. In: CVPR
  (2025)

\bibitem{zhang2024view}
Zhang, Q., Wang, L., Patel, V.M., Xie, X., Lai, J.: View-decoupled transformer
  for person re-identification under aerial-ground camera network. In: CVPR.
  pp. 22000--22009 (2024)

\bibitem{zhao2024clip}
Zhao, H., Qi, L., Geng, X.: Clip-dfgs: A hard sample mining method for clip in
  generalizable person re-identification. ACM T MULTIM COMP  \textbf{21}(1),
  1--20 (2024)

\bibitem{zhao2025cilp}
Zhao, H., Qi, L., Geng, X.: Cilp-fgdi: Exploiting vision-language model for
  generalizable person re-identification. IEEE TIFS  (2025)

\bibitem{Zhao2021M3L}
Zhao, Y., Zhang, J., et~al.: Learning to generalize unseen domains via
  memory-based multi-source meta-learning for person re-identification. In:
  CVPR (2021)

\bibitem{zheng2021deep}
Zheng, Z., Zheng, Z., Zheng, W.S., Tao, D.: Deep smal-based 3d reconstruction
  for animal re-identification. In: ACM MM. pp. 4680--4688 (2021)

\bibitem{zhu2024seas}
Zhu, H., Budhwant, P., Zheng, Z., Nevatia, R.: Seas: Shape-aligned supervision
  for person re-identification. In: CVPR. pp. 164--174 (2024)

\bibitem{zou2020joint}
Zou, Y., Yang, X., Yu, Z., Kumar, B.V., Kautz, J.: Joint disentangling and
  adaptation for cross-domain person re-identification. In: ECCV. pp. 87--104.
  Springer (2020)

\end{thebibliography}
}


\end{document}